\documentclass[twoside]{article}
\usepackage[accepted]{aistats2025}
\usepackage{amssymb}
\usepackage{amsmath}
\usepackage{amsbsy}
\usepackage{amsfonts}
\usepackage{breqn}
\usepackage{mathtools}
\usepackage{graphicx}
\usepackage{dblfloatfix}
\usepackage{subcaption}
\usepackage{svg}
\usepackage{tikz}
\usepackage{pgfplots}
\usepackage{float}
\usepackage{hyperref}
\usepackage{booktabs}
\usepackage{microtype}
\usepackage{multirow}
\usepackage{multicol}
\usepackage{adjustbox}
\usepackage{booktabs}
\usepackage{cuted}
\usepackage{placeins}
\usepackage{natbib}
\allowdisplaybreaks
\usetikzlibrary{shapes, arrows, calc, positioning, fit, decorations.pathreplacing}
\tikzstyle{encoder} = [trapezium, draw, fill=green!20, trapezium angle=85, shape border rotate = 270, minimum width=2.2em]
\tikzstyle{encoder_b} = [trapezium, draw, fill=green!20, trapezium angle=85, shape border rotate = 90, minimum width=2.2em]
\tikzstyle{decoder} = [trapezium, draw, fill=blue!20, trapezium angle=85, shape border rotate = 90, minimum width=2.2em]
\tikzstyle{function_estimator} = [draw, fill=orange!20, minimum height=2.5em, minimum width=2.5em]
\tikzstyle{input} = [coordinate]
\tikzstyle{output} = [coordinate]
\tikzstyle{mynode_blue}=[thick, draw, fill=blue!20, circle, minimum size=28]
\tikzstyle{mynode_green}=[thick, draw, fill=green!20, circle, minimum size=28]
\pgfplotsset{compat=1.16}

\begin{document}
\runningauthor{Yang, Feng, Wang, Leonard, Dieng, Allen-Blanchette}
\twocolumn[
\aistatstitle{Behavior-Inspired Neural Networks for Relational Inference}
\aistatsauthor{Yulong Yang$^{*,\dagger}$ \And Bowen Feng$^{*}$ \And Keqin Wang}
\aistatsauthor{Naomi Ehrich Leonard \And Adji Bousso Dieng \And Christine Allen-Blanchette$^{\dagger}$}
\aistatsaddress{Princeton University} ]
\begin{abstract}
    From pedestrians to Kuramoto oscillators, interactions between agents govern how dynamical systems evolve in space and time. Discovering how these agents relate to each other has the potential to improve our understanding of the often complex dynamics that underlie these systems. Recent works learn to categorize relationships between agents based on observations of their physical behavior. These approaches model relationship categories as outcomes of a categorical distribution which is limiting and contrary to real-world systems, where relationship categories often intermingle and interact. In this work, we introduce a level of abstraction between the observable behavior of agents and the latent categories that determine their behavior. To do this, we learn a mapping from agent observations to agent preferences for a set of latent categories. The learned preferences and inter-agent proximity are integrated in a nonlinear opinion dynamics model, which allows us to naturally identify mutually exclusive categories, predict an agent's evolution in time, and control an agent's behavior. Through extensive experiments, we demonstrate the utility of our model for learning interpretable categories, and the efficacy of our model for long-horizon trajectory prediction.
\end{abstract}
%
\section{Introduction}
\begin{figure*}[t]
    \centering
    \includegraphics[width=0.85\textwidth]{Images/BINN_AIStats_Cartoon.pdf}
    \label{Fig:DLC_Cartoon}
    \caption{\textbf{Behavior-inspired neural network for relational inference.} BINNs allow for relational inference using inter-agent distance and a learned representation of inter-category interactions. The physical states of agents are encoded to preferences for a set of latent categories. Preferences are propagated forward in time using a nonlinear opinion dynamics model with learned parameters. Predicted preferences are then decoded to physical states as predicted future physical states.}
    \vskip -0.1in
\end{figure*}
Multi-agent systems are found in domains as diverse as astronomy~\citep{villanueva2022inferring, lemos2023rediscovering}, biology~\citep{fiorelli2006multi, seeley2012bee, young2013starling}, physics~\citep{gull2013superconductivity, browaeys2020many}, and sports~\citep{hauri2021multi, wang2024tacticai}.
Understanding how these systems evolve in time has the potential to provide insights useful for the discovery of unknown physics, the rules governing collective behavior, and engineering design.
Predicting the evolution of complex systems is a fundamental challenge in the learning literature.
Early black-box models determine future states from past states without regard for contextual information~\citep{hochreiter1997long,sutskever2014sequence}.
More recent approaches improve upon these models by incorporating information about environmental conditions and the behavior of other agents~\citep{alahi2016social, gupta2018social, casas2018intentnet, deo2018convolutional, sadeghian2019sophie}.
While incorporating contextual information has led to more powerful trajectory prediction models, the opacity of these models limits our ability to leverage them to better understand the role of inter-agent relationships. 

Recent work in relational inference attempts to address this limitation by explicitly modeling inter-agent relationships~\citep{sukhbaatar2016learning,santoro2017simple,kipf2018neural, graber2020dynamic, xu2022groupnet, xu2023eqmotion}.
Discovering inter-agent relationships is challenging, however, since ground truth labels are typically unavailable, and the relevant relationships may be unknown at design time.
Graph neural network (GNN) based approaches such as ~\citet{kipf2018neural, graber2020dynamic, xu2022groupnet, xu2023eqmotion} model inter-agent relationship categories with a categorical variable.
In their seminal work, the authors of \citet{kipf2018neural} learn a latent representation of inter-agent relationships in a trajectory prediction pipeline. 
The authors of \citet{graber2020dynamic} improve on this approach by modeling inter-agent relationships as time varying and the authors of \citet{xu2022groupnet} improve model expressivity by representing node features as a hypergraph.
While these methods are able to predict future observations of a number of systems, the underlying assumption that inter-agent relationships are determined by a single relationship category diverges from what we observe in the real world.

In contrast to these models, opinion dynamics models assume agents have preferences for multiple categories, and that these preferences evolve in time.
The nonlinear opinion dynamics (NOD) model introduced in~\citet{leonard2024fast}, has been used to model a variety of societal systems~\citep{das2014modeling, rossi2020opinion, leonard2021nonlinear, franci2021analysis, cathcart2022opinion, bizyaeva2022switching, bizyaeva2024active, hu2023emergent, ordorica2024opinion, wang2025understanding}. 
The nonlinear nature of this model introduces a bifurcation which allows preferences to quickly and flexibly respond to changes in environmental inputs. 
A limitation of this method, however, is that the relationship between agent preferences for a set of categories and their observable behavior must be known a priori.

In our model, \textit{Behavior-Inspired Neural Network} (BINN), we combine the flexibility of GNN models with the interpretability of nonlinear opinion dynamics for a new approach to relational inference. 
Concretely, our contributions are the following:
\vspace{-0.1in}
\begin{itemize}
    \itemsep0em 
    \item In contrast to existing opinion dynamics models which require observability of agent preferences, we learn a representation of agent preferences from physical observations.
    \item In contrast to existing relational inference approaches which model relationship categories as a categorical variable, we use the NOD inductive bias to model the evolution of real-valued agent preferences on a set of latent categories.
    \item By incorporating the NOD inductive bias, we can identify mutually exclusive categories, predict an agent's evolution in time, and control an agent's behavior.
\end{itemize}
We demonstrate the utility of our approach for identifying mutually exclusive categories on multiple illustrative examples and demonstrate the efficacy of our approach for trajectory prediction tasks.
%
%
\section{Related Work}
%
%
\textbf{Trajectory prediction.} 
In contrast to traditional control approaches which tune a set of model parameters from all available trajectories~\citep{brunton2016discovering, galioto2020bayesian, paredes2021identification, paredes2024experimental, paredes2024output, richards2024output}, early deep learning approaches learn to map a sequence of input states to future states directly~\citep{sutskever2014sequence}. 
Later efforts incorporated external influences (e.g., behavior of other agents, environmental conditions)~\citep{alahi2016social, gupta2018social, casas2018intentnet, deo2018convolutional, sadeghian2019sophie},
physical priors~\citep{greydanus2019hamiltonian, mason2022learning, mason2023learning, allen2024hamiltonian},
and temporal dependencies~\citep{vemula2018social, sadeghian2019sophie, mangalam2020not, yuan2021agentformer, giuliari2021transformer, rezaei2024alternators}, and
recent work has used graph based methods to model multi-agent dynamics~\citep{yu2020spatio, gao2020vectornet}.
While these methods can predict future system states, 
%
%
they do not predict which agents interact or how they interact, a limitation which motivates relational inference.
%

\textbf{Relational inference.}
The goal of relational inference is to infer inter-agent relationships in a multi-agent system.
This task is challenging since, in general, the relationships between individual agents are unobservable. 
Early works~\citep{sukhbaatar2016learning, foerster2016learning} focus on learning the communication protocols for multi-agent systems.
Neural Relational Inference~\citep{kipf2018neural} proposed a variational autoencoding~\citep{kingma2013auto} graph neural network framework for learning time-invariant relationships between individual agents.
Further developments focused on increasing the expressivity and applicability of this framework by incorporating factorized graphs to model different types of interactions~\citep{webb2019factorised}; dynamic encoders to model time-varying inter-agent relations~\cite{graber2020dynamic}; edge-to-edge message passing for more efficient information sharing~\citep{chen2021neural}; shared decoder conditioned on edge types to learn a mapping for future states~\citep{lowe2022amortized}; hypergraphs to accommodate interactions of different spatial scales~\citep{xu2022groupnet}; and Euclidean transformation equivariance to improve generalization to varying scenes~\cite{xu2023eqmotion}.
Other works leverage different architectures to infer inter-agent relations such as meta-learning to map inputs to edge and node values~\citep{alet2019neural}; reservoir computing to increase the efficiency of relational inference~\citep{wang2023effective}; partial correlates of latent representations to infer connections~\citep{wang2024structural}; and diffusion models to reconstruct missing connections~\citep{zheng2024diffusion}.
Our work differs from these approaches in our representation of relationship categories.
Our categories exist in a space of agent preferences rather than the physical space, and our categories are flexible and interacting, rather than mutually exclusive and independent.


\textbf{Consensus dynamics.}
In control and robotics, consensus dynamics~\citep{bullo2018lectures} have been used in a myriad of settings to model the dynamics and control the behavior of multi-agent systems.
In~\citet{levine2000self}, the authors propose a linear model for prediction and control of multi-agent systems.
In~\citet{leonard2010coordinated}, the authors use a linear consensus dynamics model for coordinated surveying with underwater gliders.
In~\citet{justh2005natural},  the authors use the consensus dynamics framework to develop rectilinear and circular consensus control laws for multi-vehicles systems.
In~\citet{leonard2007collective},  the authors use consensus dynamics to improve data collection in mobile sensor networks.
In~\citet{ballerini2008interaction}, the authors use consensus dynamics to understand the robustness of starling flocking behavior.
Even with this breadth of application, there are drawbacks to a linear model of opinion formation; 
specifically, the naive implementation results in dynamics that only yields consensus~\citep{altafini2012consensus, dandekar2013biased} and the formation of opinions in response to inputs is slow.
%

\textbf{Nonlinear opinion dynamics.} 
The noted short comings of linear consensus dynamics models are resolved in the nonlinear opinion dynamics model proposed in~\citet{leonard2024fast}. 
As the model is nonlinear, the dynamics result in a bifurcation and opinions can evolve to dissensus quickly.
Nonlinear opinion dynamics have been used to model a variety of systems. In~\citet{das2014modeling, rossi2020opinion, leonard2021nonlinear, franci2021analysis} nonlinear opinion dynamics model information spread in settings such as political polarisation.
In~\citet{hu2023emergent}, nonlinear opinion dynamics is used to resolve deadlock, and
in~\citet{cathcart2022opinion}, it is used for collision avoidance in human-robotic systems.
%
%
In~\citet{bizyaeva2024active, ordorica2024opinion}, nonlinear opinion dynamics is used for modeling risk in epidemic models.
In contrast to these works, where the model has direct access agent opinions (i.e., preferences), we learn a mapping between physical states and agent preferences.
%
\section{Background}\label{sec:background}
Nonlinear dynamics differ from linear dynamics in that they are able to exhibit bifurcations~\citep{golubitsky2012singularities, guckenheimer2013nonlinear, strogatz2018nonlinear}.
Nonlinear opinion dynamics~\citep{bizyaeva2022nonlinear} leverage bifurcations for fast and flexible decision making even with weak input signals~\citep{leonard2024fast}.
Our BINN model integrates a nonlinear opinion dynamics inductive bias to interpretably and flexibly model agent behavior.
%
%
\subsection{Nonlinear opinion dynamics}
\label{subsec:nonlinear_opinion_dynamics}
Consider a multi-agent system of $\mathcal{N}_{\mathrm{a}}\in\mathbb{N}$ agents, each possessing real-valued preferences about $\mathcal{N}_{\mathrm{o}}\in\mathbb{N}$ categories.
Each category represents a belief or desire (e.g., take a physical action, undertake a task), and
an agent's preference for a category can be positive, neutral, or negative 
(the preference magnitude corresponds to preference strength).
Changes to an agent's preference for a category depend on extrinsic and intrinsic parameters.
Concretely, the changes to agent $i$'s preference for category $j$ can be determined by the differential equation proposed in~\citet{leonard2024fast}
\begin{multline}
    \dot{z}_{ij} = -d_{ij}z_{ij} + \mathcal{S}\Bigg(u_{i}\Bigg(\alpha_{ij}z_{ij} + \sum_{\substack{k=1\\ k\neq i}}^{\mathcal{N}_{\mathrm{a}}}a^{\mathrm{a}}_{ik}z_{kj}\\
    + \sum_{\substack{l=1\\ l\neq j}}^{\mathcal{N}_{\mathrm{o}}}a^{\mathrm{o}}_{jl}z_{il} + \sum_{\substack{k=1\\ k\neq i}}^{\mathcal{N}_{\mathrm{a}}}\sum_{\substack{l=1\\ l\neq j}}^{\mathcal{N}_{\mathrm{o}}} a^{\mathrm{a}}_{ik}a^{\mathrm{o}}_{jl}z_{kl}\Bigg)\Bigg) + b_{ij}. \label{eqn:nonlinear_opinion_dynamics}
\end{multline}

The parameters $d_{ij}\ge 0$, $u_i\ge 0$, and $\alpha_{ij}\ge 0$ are intrinsic to the agent.
The damping $d_{ij}$ represents how resistant agent $i$ is to forming a preference for category $j$,
the attention $u_{i}$ represents how attentive agent $i$ is to the preferences of other agents, and the self-reinforcement $\alpha_{ij}$ represents how resistant agent $i$ is to changing its preference about category $j$.

The parameters $a^a_{jk}$, $a^o_{jl}$, and $b_{ij}$ are extrinsic to the agent.
The communication matrix $[a^a_{ik}]$ describes the communication strength between agent $i$ and agent $k$,
the belief matrix $[a^o_{jl}]$ describes the correlation of preferences for category $j$ and category $l$, and
the environmental input $b_{ij}$ describes the impact of the environment on agent $i$'s preference for category $j$.
The saturating function $\mathcal{S}$ is selected such that  $S\left(0\right)=0$, $S'\left(0\right)=1$, and $S'''\left(0\right)\neq0$~\citep{leonard2024fast}. 
We use $\operatorname{tanh}\left(\cdot\right)$ for the saturation function in our experiments.

For a given communication matrix $[a^o_{jl}]$ and belief matrix $[a^o_{jl}]$, the sensitivity of preference formation is determined by an agent's intrinsic parameters $d_{ij}$, $u_i$, and $\alpha_{ij}$, and the equilibrium value of each preference is a function of the environmental input $b_{ij}$ (see Figure \ref{fig:nonlinear_opinion_dynamics_bifurcation}).
%
%
\begin{figure*}[t]
    \centering
    \resizebox{0.95\textwidth}{!}{
    \begin{tikzpicture}[>={stealth'}, line width = 0.25mm]

    \draw[->] (-1,-1)--(3,-1) node [right] {\tiny $b_{ij}$};
    \draw (1,-1)--(1,-0.9);
    \node [below] at (1,-1) {\tiny $0$};
    \node [below] at (1,-1.4) {\tiny (c) $u_i > u^*$};
    \draw[->] (-1,-1)--(-1,1) node [left] {\tiny $z_{ij}$};
    \draw (-1,0)--(-0.9,0);
    \node [left] at (-1,0) {\tiny $0$};

    \draw[black] plot[variable=\y,domain=-0.8:-0.5,smooth] ({(4*\y*\y*\y*\y*\y*\y*\y + 1.2*\y*\y*\y*\y*\y + 2*\y*\y*\y - 1.8*\y)*2 + 1}, {\y});
    \draw[dotted] plot[variable=\y,domain=-0.5:0.5,smooth] ({(4*\y*\y*\y*\y*\y*\y*\y + 1.2*\y*\y*\y*\y*\y + 2*\y*\y*\y - 1.8*\y)*2 + 1}, {\y});
    \draw[black] plot[variable=\y,domain=0.5:0.8,smooth] ({(4*\y*\y*\y*\y*\y*\y*\y + 1.2*\y*\y*\y*\y*\y + 2*\y*\y*\y - 1.8*\y)*2 + 1}, {\y});

    \draw[->] (-7,-1)--(-3,-1) node [right] {\tiny $b_{ij}$};
    \draw (-5,-1)--(-5,-0.9);
    \node [below] at (-5,-1) {\tiny $0$};
    \node [below] at (-5,-1.4) {\tiny (b) $u_i < u^*$};
    \draw[->] (-7,-1)--(-7,1) node [left] {\tiny $z_{ij}$};
    \draw (-7,0)--(-6.9,0);
    \node [left] at (-7,0) {\tiny $0$};

    \draw[color=black, domain=-6.5:-3.5,smooth] plot (\x, {0.8*tanh(0.85*(\x+5))});

    \draw[->] (-13,-1)--(-9,-1) node [right] {\tiny $u_{i}$};
    \draw (-11,-1)--(-11,-0.9);
    \node [below] at (-11,-1) {\tiny $u^{*}$};
    \node [below] at (-11,-1.4) {\tiny (a)};
    \draw[->] (-13,-1)--(-13,1) node [left] {\tiny $z_{ij}$};
    \draw (-13,0)--(-12.9,0);
    \node [left] at (-13,0) {\tiny $0$};

    \draw (-13,0)--(-11,0);
    \draw[dotted] (-11,0)--(-9,0);
    \draw[color=black, domain=-11:-9, samples=150] plot (\x, {0.07*sqrt(50*(\x+11))});
    \draw[color=black, domain=-11:-9, samples=150] plot (\x, {-0.07*sqrt(50*(\x+11))});
    
    \end{tikzpicture}
    }
    \vskip -0.05in
    \caption{\textbf{Sensitivity to environmental inputs $b_{ij}$.} Solid lines represent stable equilibria and dotted line represents unstable equilibria. \textbf{(a)} We show the pitchfork bifurcation characteristic to Equation~\eqref{eqn:nonlinear_opinion_dynamics}. The number, location, and stability of equilibria changes with the attention parameter $u_{i}$. \textbf{(b)} For attention $u_{i}< u^{*}$ preferences change linearly with environmental inputs. \textbf{(c)} For attention $u_{i}> u^{*}$ preferences change rapidly in response to environmental inputs with hysteresis encoding memory of previous states.}
    \vskip -0.1in
    \label{fig:nonlinear_opinion_dynamics_bifurcation}
\end{figure*}
%
%
\begin{figure*}[!b]
    \centering
    \resizebox{1.4\columnwidth}{!}{
    \begin{tikzpicture}[>={stealth'}, line width = 0.25mm]

    \node [input, name=input] {};
    \node [input, name=input_draw, right = 0.5cm of input] {};
    \node [encoder, right = 1cm of input] (encoder_1) {$E_{z}$};
    \node [encoder, below = 0.32cm of encoder_1] (encoder_b_0) {$E_{b}$};
    \node [right = 0.5cm of encoder_1] (z_agt_t_1) {$\mathbf{z}_{i, t}$};
    \node (b_agt_0) at (z_agt_t_1 |- encoder_b_0) {$\mathbf{b}_{i, t}$};
    \node [function_estimator, right = 0.9cm of z_agt_t_1] (function_1) {$f_{\mathrm{NOD}}$};
    \node [right = 0.75cm of function_1] (z_agt_dot_1) {$\dot{\mathbf{z}}_{i, t}$};
    \node [right = 0.75cm of z_agt_dot_1] (z_agt_t1_1) {$\mathbf{z}_{i, t+1}$};
    \node [decoder, right = 0.5cm of z_agt_t1_1] (decoder_1) {$D_{x}$};
    \node [right = 0.75cm of decoder_1] (x_pred_1) {$\hat{\mathbf{x}}_{i, t+1}$};
    \node [encoder_b, below = 0.32cm of decoder_1] (encoder_b_1) {$E_{b}$};
    \node (b_agt_1) at (z_agt_t1_1 |- encoder_b_1) {$\mathbf{b}_{i, t+1}$};

    \draw [draw,->] (input) -- node [name=u]{} node [very near start, left=0.12cm] (input_label){$\mathbf{x}_{i, t}$} (encoder_1);
    \draw [draw,->] (input_draw) |- (encoder_b_0);
    \draw [draw,->] (encoder_1) -- (z_agt_t_1);
    \draw [draw,->] (encoder_b_0) -- (b_agt_0);
    \draw [draw,->] (b_agt_0.east) -- +(0.4, 0cm) |- (function_1.west);
    \draw [draw,->] (z_agt_t_1) -- (function_1);
    \draw [draw,->] (function_1) -- (z_agt_dot_1);
    \draw [draw,->] (z_agt_dot_1) -- (z_agt_t1_1);
    \draw [draw,->] (z_agt_t1_1) -- (decoder_1);
    \draw [draw,->] (decoder_1) -- (x_pred_1);
    \draw [draw,->] (x_pred_1) |- (encoder_b_1);
    \draw [draw,->] (encoder_b_1) -- (b_agt_1);

    \node [function_estimator, below = 1.2cm of function_1] (function_2) {$f_{\mathrm{NOD}}$};
    \node (z_agt_dot_2) at (z_agt_dot_1 |- function_2) {$\dot{\mathbf{z}}_{i, t+1}$};
    \node (z_agt_t1_2) at (z_agt_t1_1 |- function_2) {$\mathbf{z}_{i, t+2}$};
    \node [decoder] (decoder_2) at (decoder_1 |- function_2) {$D_{x}$};
    \node (x_pred_2) at (x_pred_1 |- function_2){$\hat{\mathbf{x}}_{i, t+2}$};

    \draw [draw,->] (function_2) -- (z_agt_dot_2);
    \draw [draw,->] (z_agt_dot_2) -- (z_agt_t1_2);
    \draw [draw,->] (z_agt_t1_2) -- (decoder_2);
    \draw [draw,->] (decoder_2) -- (x_pred_2);
    \draw [draw,->] (b_agt_1) -| (function_2);
    \draw [draw,->] (z_agt_t1_1.south) -- +(0, -0.35cm) -|(function_2.north);
    
    \end{tikzpicture}
    }
    \caption{\textbf{Behavior-inspired neural network architecture overview.} Our network takes agent states as inputs and outputs predicted next states, while maintaining a representation of agent preferences for a set of latent categories. The network $E_{z}$ encodes physical states to preferences for a set of latent categories and $E_{b}$ encodes physical states into latent environmental inputs. In the latent space, we compute future preferences using the nonlinear opinion dynamics block $f_{\mathrm{NOD}}$, and use the decoder $D_{x}$ to map predicted preferences to predicted physical states. The latent dynamics are unrolled for multi-step trajectory prediction.}
    \label{fig:BINN_architecture}
\end{figure*}

\textbf{Mutually exclusive categories.}
In the nonlinear opinion dynamics framework, the presence of mutually exclusive categories simplifies the model. For example, consider the case of $\mathcal{N}_a$ agents and $\mathcal{N}_o=2$ categories. The categories are said to be mutually exclusive if $a^{o}_{12},a^{o}_{21}\le 0$ and $z_{i1}=-cz_{i2}$ (i.e. a positive preference for one category implies a negative preference for the other).
In this setting, the dynamics for $z_{i1}$ and $z_{i2}$ decouple, and
Equation \eqref{eqn:nonlinear_opinion_dynamics} reduces to
\begin{equation}
    \resizebox{0.91\columnwidth}{!}{
    $\dot{z}_{i} = -d_{i}z_{i} + \mathcal{S}\left(u_{i}\left(\tilde{\alpha}_{i}z_{i} + \sum_{\substack{k=1\\ k\neq i}}^{\mathcal{N}_{\mathrm{a}}}\tilde{a}_{ik}z_{kj}\right)\right) + b_{i}. \label{eqn:nonlinear_opinion_dynamics_mutually_exclusive_reduced}$
    }
\end{equation}
In our model, we leverage the presence of mutually exclusive categories to systematically identify opportunities for dimensionality reduction. 
Additional details are provided in Appendix~\ref{apx:experiments_additional_mutually_exclusive}. 
%
%
\section{Method}\label{sec:method}
In this section we present our BINN model for relational inference (see Figure \ref{fig:BINN_architecture}). 
Given trajectories of a multi-agent system, our goal is to predict agent behavior in an interpretable way by inferring the intrinsic and extrinsic characteristics of agents as described by the nonlinear opinion dynamics model (see Section~\ref{sec:background} and Appendix~\ref{apx:nonlinear_opinion_dynamics}).

We define a multi-agent system as a system of $\mathcal{N}_{\mathrm{a}}\in\mathbb{N}$ agents, each with real-valued preferences for a set of $\mathcal{N}_{\mathrm{o}}\in\mathbb{N}$ categories.
We learn a mapping between the observed behavior of agents and their preferences for a set of latent categories using a graph neural network, and the parameters of a nonlinear opinion dynamics model which determine the evolution of agent preferences.

We model the evolution of agent preference by the dynamical equation presented in Equation~\eqref{eqn:nonlinear_opinion_dynamics}.
With this formulation, we can control agent preferences by varying the environmental input as shown in Figure~\ref{fig:pendulum_bifurcation}.

We train our model using $N$ trajectories of $T$ observations, where each observation has dimension $d$.
For a given trajectory, we denote the observed state of agent $i$ at time $t$ by the concatenation of position and velocity $\mathbf{x}_{i,t}=\left[\mathbf{p}_{i,t}, \mathbf{v}_{i,t}\right]\in\mathbb{R}^{d}$, and the preferences of agent $i$ at time $t$ by $\mathbf{z}_{i,t}\in\mathbb{R}^{\mathcal{N}_{o}}$.
%
%
\subsection{Encoder}\label{subsec:encoder}
We use separate encoding networks to learn the mappings from agent states to agent preferences and from agent states to environmental inputs. 
We refer to the first of these encoders as the preference encoder $E_{z}$, and the latter as the environmental input encoder $E_{b}$.
Each encoder takes state observations of the multi-agent system at time $t$ as input and processes them in an message passing neural network (MPNN)~\citep{gilmer2017neural}. 
The multi-agent system is represented as a fully-connected graph with node values determined by the physical state of each agent. 

Our preference encoder $E_{z}$ performs the following message passing functions for agent $i$ at timestep $t$:
\begin{alignat}{2}
    && \quad\quad \mathbf{z}'_{i,t}&= f_{\mathrm{emb}}^{{z}}\left(\mathbf{x}_{i,t}\right),\\
    v\rightarrow e:&& \quad\quad \mathbf{m}^{z}_{\left(i, k\right),t}&= f_{v\rightarrow e}^{{z}}\left(\mathbf{z}'_{i,t}, \mathbf{z}'_{k,t}\right),\\
    e\rightarrow v:&& \quad\quad \mathbf{z}_{i,t}&= f_{e\rightarrow v}^{{z}}\left(\textstyle\sum_{k\neq i} \mathbf{m}^{z}_{ \left(i, k\right),t}, \mathbf{z}'_{i,t}\right),
\end{alignat}
where $f_{\mathrm{emb}}^{z}$, $f_{v\rightarrow e}^{z}$, and $f_{e\rightarrow v}^{z}$ are 3-layer MLPs.
The environmental input encoder $E_b$ is designed similarly, and performs the following message passing functions for agent $i$ at timestep $t$:
\begin{alignat}{2}
    && \quad\quad \mathbf{b}'_{i,t}&= f_{\mathrm{emb}}^{{b}}\left(\mathbf{x}_{i,t}\right),\\
    v\rightarrow e:&& \quad\quad \mathbf{m}^{b}_{\left(i, k\right),t}&= f_{v\rightarrow e}^{{b}}\left(\mathbf{b}'_{i,t}, \mathbf{b}'_{k,t}\right),\\
    e\rightarrow v:&& \quad\quad \mathbf{b}_{i,t}&= f_{e\rightarrow v}^{{b}}\left(\textstyle\sum_{k\neq i} \mathbf{m}^{b}_{ \left(i, k\right),t}, \mathbf{b}'_{i,t}\right),
\end{alignat}
where $f_{\mathrm{emb}}^{{b}}$, $f_{v\rightarrow e}^{b}$, and $f_{e\rightarrow v}^{b}$ are 3-layer MLPs.
%
%
\subsection{Latent nonlinear opinion dynamics}\label{subsec:latent_nonlinear_opinion_dynamics}
We use the nonlinear opinion dynamics formulation in Equation \eqref{eqn:nonlinear_opinion_dynamics}  to model the evolution of agent preferences on a set of latent categories,
\begin{equation}
    \dot{\mathbf{z}}_{i, t+1} = f_{\mathrm{NOD}}\left(\mathbf{z}_{i, t}, \mathbf{b}_{t}, \mathbf{A}^{\mathrm{a}}_{t}\right).
\end{equation}
We learn the intrinsic agent parameters, $\mathbf{d}$, $\mathbf{u}$, and $\boldsymbol{\alpha}$, the extrinsic belief matrix $\mathbf{A}^{\mathrm{o}}$, and compute the inter-agent communication matrix $\mathbf{A}^{\mathrm{a}}_t$ at timestep $t$.

We compute future preferences using Euler integration,
\begin{equation}
    \mathbf{z}_{i, t+1} = \mathbf{z}_{i, t} + f_{\mathrm{NOD}}\left(\mathbf{z}_{i, t}, \mathbf{b}_{t}, \mathbf{A}^{\mathrm{a}}_{t}\right)\Delta t,
\end{equation}
where the timestep $\Delta t$ is dataset dependent.
%
%

\subsection{Communication matrix}\label{subsec:communication_matrix}
We define the communication matrix $\mathbf{A}^{\mathrm{a}}_{t}$ as a function of inter-agent proximity. In our baseline BINN model, we define $\mathbf{A}^{\mathrm{a}}_{t}$ as the distance between observed positions, 
\begin{equation}
    a^{\mathrm{a}}_{(i,j),t} = \lVert\mathbf{p}_{i,t} - \mathbf{p}_{j,t}\rVert^{2},\label{eqn:communication_matrix_baseline}
\end{equation}
We also introduce \textit{Behavior-Inspired Neural Network\raisebox{+0.2ex}{+}} (BINN\raisebox{+0.2ex}{+}), where we incorporate an informed hypothesis of how agents in the system communicate. 
For example, for systems in which agents have more influence on each other when they are closer (e.g., human interaction) we define $\mathbf{A}^{\mathrm{a}}_{t}$ as the inverse distance between observed positions,
\begin{equation}
    a^{\mathrm{a}}_{(i,j),t} = \frac{1}{\lVert\mathbf{p}_{i,t} - \mathbf{p}_{j,t}\rVert^{2} + \epsilon},
\end{equation}
where $\epsilon$ is a small constant.
%
%
\subsection{Decoder}
Our decoding network $D_x$, is an MPNN that maps agent preferences to predictions of agent states.
At every timestep $t$, the latent preferences of the multi-agent system are represented as a fully-connected graph with node values determined by the preferences of each agent.
%

Our decoder $D_{x}$ performs the following message passing functions for agent $i$ at timestep $t$:
\begin{alignat}{2}
    && \quad\quad\hspace{-2mm} \hat{\mathbf{x}}'_{i,t} &= f^{x}_{\mathrm{dec}}\left(\mathbf{z}_{i,t}\right),\\
    v\rightarrow e:&& \quad\quad\hspace{-2mm} \mathbf{m}^{x}_{\left(i, k\right), t} &= f_{v\rightarrow e}^{{x}}\left(\hat{\mathbf{x}}'_{i, t}, \hat{\mathbf{x}}'_{k, t}\right),\\
    e\rightarrow v:&& \quad\quad\hspace{-2mm} \hat{\mathbf{x}}_{i, t} &= f_{e\rightarrow v}^{{x}}\left(\textstyle\sum_{k\neq i} \mathbf{m}^{x}_{\left(i, k\right), t}, \hat{\mathbf{x}}'_{i,t}\right),
\end{alignat}
where $f_{\mathrm{dec}}^{{x}}$, $f_{v\rightarrow e}^{x}$, and $f_{{e\rightarrow v}}^{{x}}$ are 3-layer MLPs.
%
%
\subsection{Loss function}
We train our model using the three component loss function,
\begin{equation}
    \mathcal{L} = \mathcal{L}_{\mathrm{pred}} + \gamma_{1}\mathcal{L}_{\mathrm{recon}} + \gamma_{2}\mathcal{L}_{\mathrm{latent}}
\end{equation}
where $\mathcal{L}_{\mathrm{pred}}$ is the prediction loss, $\mathcal{L}_{\mathrm{recon}}$ is the reconstruction loss and $\mathcal{L}_{\mathrm{latent}}$ is the latent loss.

The prediction loss, $\mathcal{L}_{\mathrm{pred}}$, is defined as the dissimilarity between the ground truth future state $\mathbf{x}_{i,t}$, and the predicted future state $\hat{\mathbf{x}}_{i,t}$,
\begin{equation}
    \mathcal{L}_{\mathrm{pred}} = \frac{1}{T\mathcal{N}_{a}}\sum_{t=1}^{T}\sum_{i=1}^{\mathcal{N}_{a}}\Big\lVert \mathbf{x}_{i,t} - \hat{\mathbf{x}}_{i,t} \Big\rVert^{2},
\end{equation}
and encourages accuracy of future state prediction. 
The reconstruction loss, $\mathcal{L}_{\mathrm{recon}}$, is defined as the dissimilarity between the ground truth initial state $\mathbf{x}_{i,0}$, and the reconstructed initial state,
\begin{equation}
    \mathcal{L}_{\mathrm{recon}} = \frac{1}{\mathcal{N}_{a}}\sum_{i=1}^{\mathcal{N}_{a}}\Big\lVert \mathbf{x}_{i,0} - \left(D_{x}\circ E_{z}\right)\left(\mathbf{x}_{i,0}\right) \Big\rVert^{2},
\end{equation}
and encourages the decoder to function as the inverse of the encoder.
The latent loss, $\mathcal{L}_{\mathrm{latent}}$, is defined as the dissimilarity of preferences predicted by the dynamical model and those encoded by the preference encoder, 
\begin{equation}
    \mathcal{L}_{\mathrm{latent}} = \frac{1}{T\mathcal{N}_{a}}\sum_{t=1}^{T}\sum_{i=1}^{\mathcal{N}_{a}}\Big\lVert \Delta\mathbf{z}_{i,t+1} - f_{\mathrm{NOD}}\left(\mathbf{z}_{i,t}\right) \Big\rVert^{2},
\end{equation}
where $\Delta \mathbf{z}_{i,t+1}=\left(\mathbf{z}_{i, t+1}-\mathbf{z}_{i, t}\right)/\Delta t$. This loss encourages alignment between the representations encoded by $E_{z}$ and those predicted by $f_{\mathrm{NOD}}$.
%
%
\section{Experiments}\label{sec:experiment}
In this section, we highlight the utility of our model for discovering interpretable representations useful for dimensionality reduction, and its efficacy for long-horizon trajectory prediction. Dataset and training details are provided in Appendices \ref{apx:datasets} and \ref{apx:training_details}.
%
%
\subsection{Interpretability of the latent space}\label{section:experiment_interpretability}
We demonstrate the interpretability of our latent representations on both the pendulum and double pendulum datasets.
%
%
\begin{figure}[t]
    \centering
    \includegraphics[width=\columnwidth]{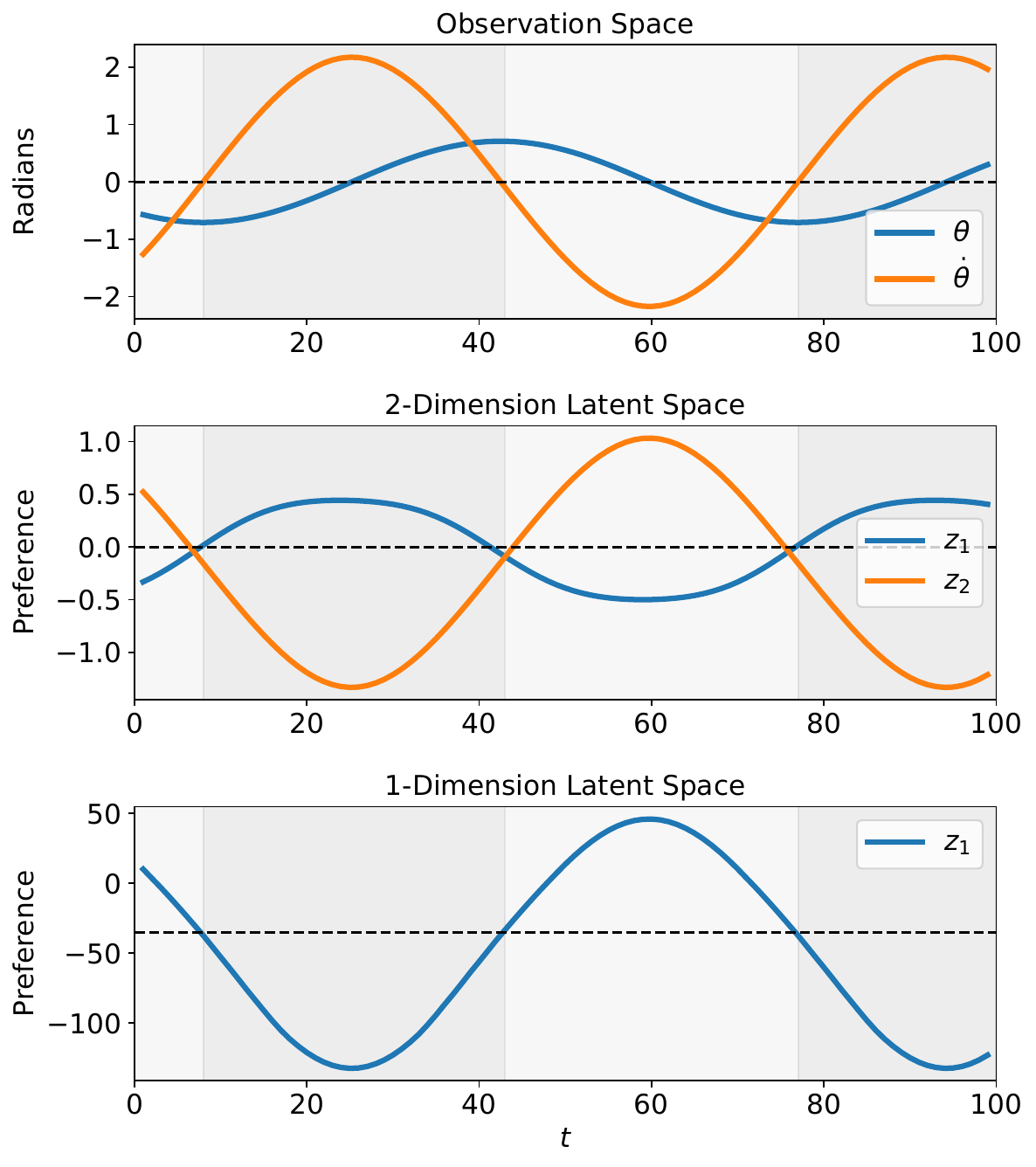}
    \caption{\textbf{Mutually exclusive pendulum preferences.} We show the observed, and latent representations of the pendulum bob. \textbf{(Top-bottom)} The observed position and velocity of the pendulum bob; the learned representation of agent preferences on a 2-dimensional latent space, and the learned representation of agent preferences on a 1-dimensional latent space. In the 2-dimensional space, preferences are out of phase indicating that they are mutually exclusive. }
    \label{fig:pendulum_2d}
    \vskip -0.1in
\end{figure}

\textbf{Pendulum.}
We model the pendulum as a single agent system represented as a graph with a single node. We construct a graphical representation at every timestep and set the node feature to the concatenation of the position and velocity of the pendulum bob.
\begin{figure}[t]
    \begin{subfigure}{\columnwidth}
        \centering
        \includegraphics[width=0.6\columnwidth]{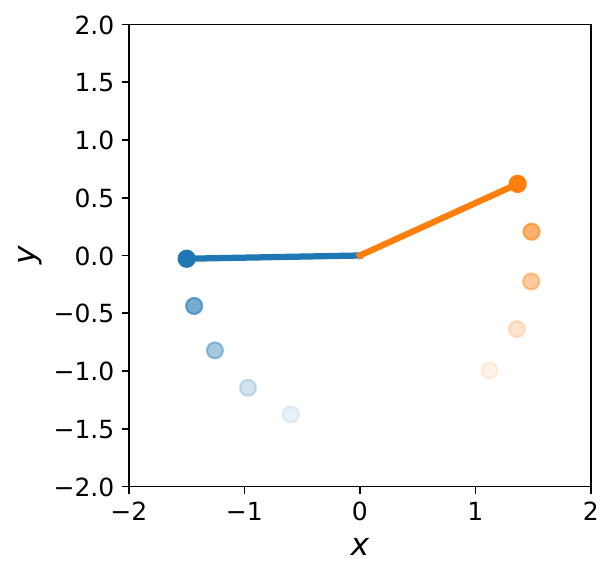}
        \caption{Example trajectories.}
        \label{subfig:pendulum_bifurcation_example_trajectory}
    \end{subfigure}
    \vskip 0.1in

    \begin{subfigure}{\columnwidth}
        \centering
        \includegraphics[width=0.9\columnwidth]{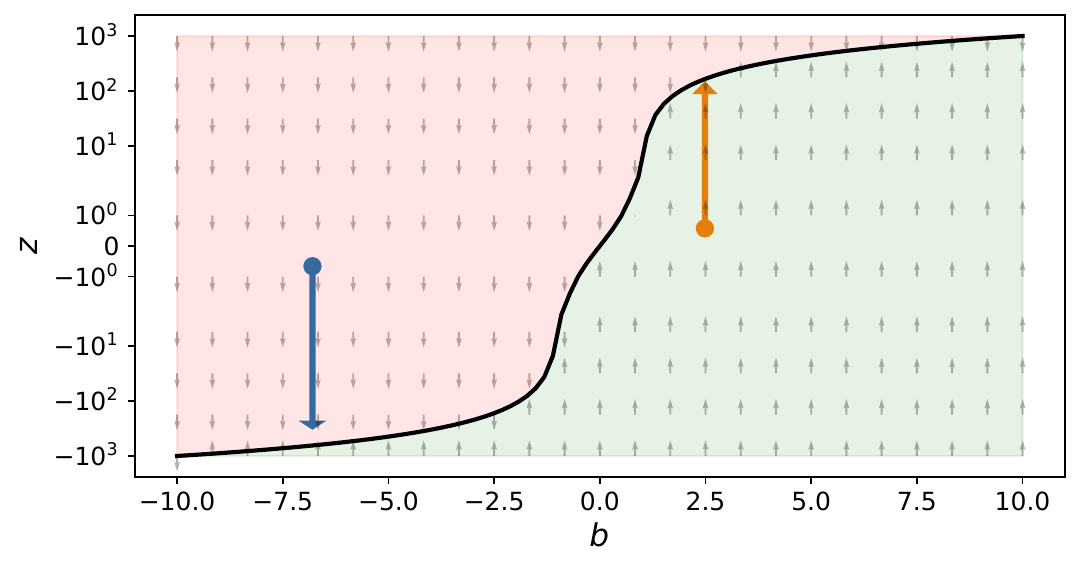}
        \caption{Bifurcation diagram.}
        \label{subfig:pendulum_bifurcation_linear_sensitivity}
    \end{subfigure}
    \caption{\textbf{Pendulum preference bifurcation diagram.} \textbf{(a)} We show a pendulum trajectory with clockwise motion (blue) and counterclockwise motion (orange). \textbf{(b)} The initial preferences are near neutral, but large environmental inputs drive preferences in opposite directions resulting in opposite motions in the physical space.}
    \vskip -0.1in
    \label{fig:pendulum_bifurcation}
\end{figure}

The relationship between observed physical states and learned preferences is shown in Figure \ref{fig:pendulum_2d}.
Our model suggests the pendulum is a 1-dimensional system which is consistent with physical understanding. 
We can see this in two ways. First, since the preferences $z_{1}$ and $z_{2}$ switch signs when the pendulum bob changes direction, we can interpret the learned categories as a desire to move clockwise ($z_{1} > 0$) and a desire to move counter-clockwise ($z_{2} > 0$).
Secondly, since the two preferences are perfectly out-of-phase and the entries of the learned belief matrix are negative,
%
%
%
\begin{equation}
\label{eq:a_o_pendulum}
    \mathbf{A}^{\mathrm{o}} = \begin{bmatrix}
        0 & -1.2947\\
        -0.9147 & 0
    \end{bmatrix},
\end{equation}
the conditions for mutual exclusivity is satisfied (see Section \ref{subsec:nonlinear_opinion_dynamics}).

We empirically validate that the dynamics of this system can be captured with a 1-dimensional latent space.
The prediction error for models with a 2-dimensional and 1-dimensional latent space are comparable i.e., $1.9718$e$-3$ MSE and $1.9916$e$-3$ MSE respectively. Qualitative results are shown in Figure \ref{fig:pendulum_2d} (bottom).

In addition to providing a mechanism for identifying opportunities for dimensionality reduction, the nonlinear opinion dynamics inductive bias also provides a mechanism for controlling preference formation through the environmental input $b_{ij}$. 
We show trajectories with different rotational directions in Figure \ref{subfig:pendulum_bifurcation_example_trajectory}, and their corresponding environmental input latent preference pairs, ($b_{ij}$, $z_{ij}$), on the bifurcation diagram in Figure \ref{subfig:pendulum_bifurcation_linear_sensitivity}.
The initial preferences for both trajectories are near neutral, but the initial environmental inputs are far from zero. The negative environmental input drives the initial preference of the blue trajectory toward an equilibrium in the preference space that corresponds to a strongly negative preference, and a clockwise motion of the pendulum bob in the physical space. 
The positive environmental input drives the initial preference of the orange trajectory toward an equilibrium in the preference space that corresponds to a strongly positive preference, and a counterclockwise motion of the pendulum bob in the physical space.
%

\textbf{Double pendulum.}
We use the double pendulum dataset to demonstrate interpretability in a more complex setting. 
\begin{figure}[t]
    \centering
    \includegraphics[width=\columnwidth]{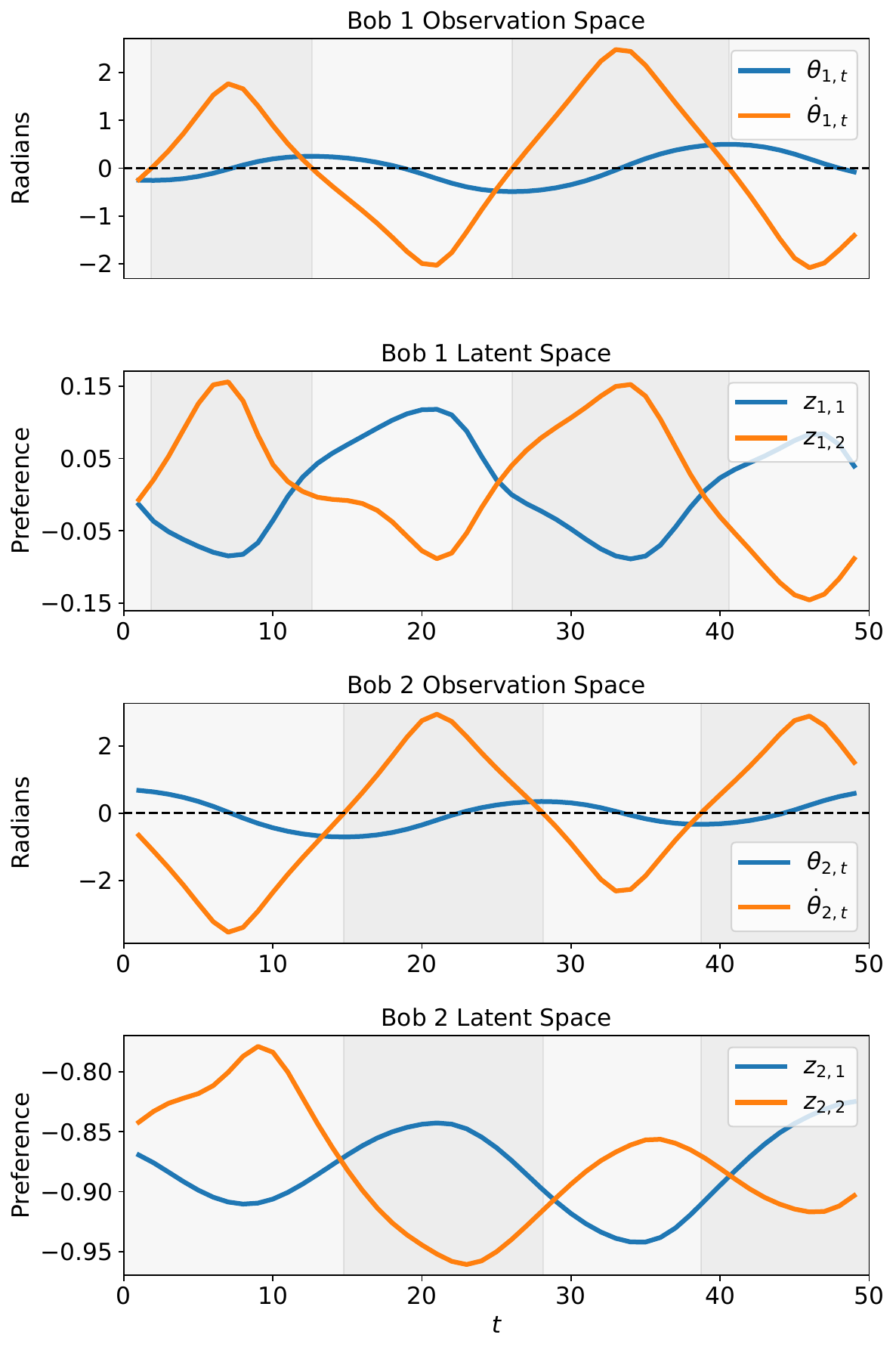}
    \caption{\textbf{Non-mutually exclusive double pendulum preferences.} We show the observed, and latent representations of each double pendulum bob. \textbf{(Top-bottom)} The observed position and velocity of the first bob; the learned preferences of the first bob; the observed position and velocity of the second bob; the learned preferences of the second bob. When the dominance of the preference switch, the physical bob's motion switches direction.}
    \vskip -0.1in
    \label{fig:double_pendulum_2d}
\end{figure}
We model the double pendulum as a two agent system represented as a graph with a two nodes. We construct a graphical representation at every timestep and set node features to the concatenation of the position and velocity of the corresponding double pendulum bob.

The relationship between the observed physical states and the learned preferences is shown in Figure \ref{fig:double_pendulum_2d}. 
For the first bob, when the preference $z_{1,1}$ switches from greater (less) than to less (greater) than preference $z_{1,2}$, the physical bob switches from clockwise (counterclockwise) to counterclockwise (clockwise) motion. The opposite is true for the second bob. 
This behavior is observed across examples (see Appendix \ref{apx:experiments_additional_double_pendulum}).
\begin{table*}[t]
    \renewcommand{\arraystretch}{1.4}
    \setlength{\tabcolsep}{10pt}
    \centering
    \caption{\textbf{Trajectory prediction.} Trajectory prediction error on the Mass-Spring, Kuramoto, TrajNet++, and NBA Player datasets is reported. Our BINN and BINN\raisebox{+0.2ex}{+} models outperform baseline models on all datasets.}
    \begin{tabular}{lcccc}
        \hline
        Network & Mass-Spring & Kuramoto & TrajNet++ & NBA Player\\
        \hline
        BINN* & $\boldsymbol{2.88\times10^{-4}}$ & $\boldsymbol{4.98\times10^{-3}}$ & $2.16\times10^{-2}$ & $5.12\times10^{-3}$\\
        BINN\raisebox{+0.2ex}{+}* & $\boldsymbol{2.88\times10^{-4}}$ & $\boldsymbol{4.98\times10^{-3}}$ & $\boldsymbol{1.20\times10^{-2}}$ & $\boldsymbol{4.19\times10^{-3}}$\\
        NRI & $3.73\times10^{-3}$ & $8.20\times10^{-3}$ & $8.49\times10^{-2}$ & -\\
        dNRI & $4.86\times10^{-3}$ & $6.41\times10^{-3}$ & $4.00\times10^{-2}$ & $6.42\times10^{-3}$\\
        \hline
    \end{tabular}
    \label{tab:experiments}
    \vskip -0.1in
\end{table*}
\begin{figure*}[!b]
    \centering
    \includegraphics[width=\textwidth]{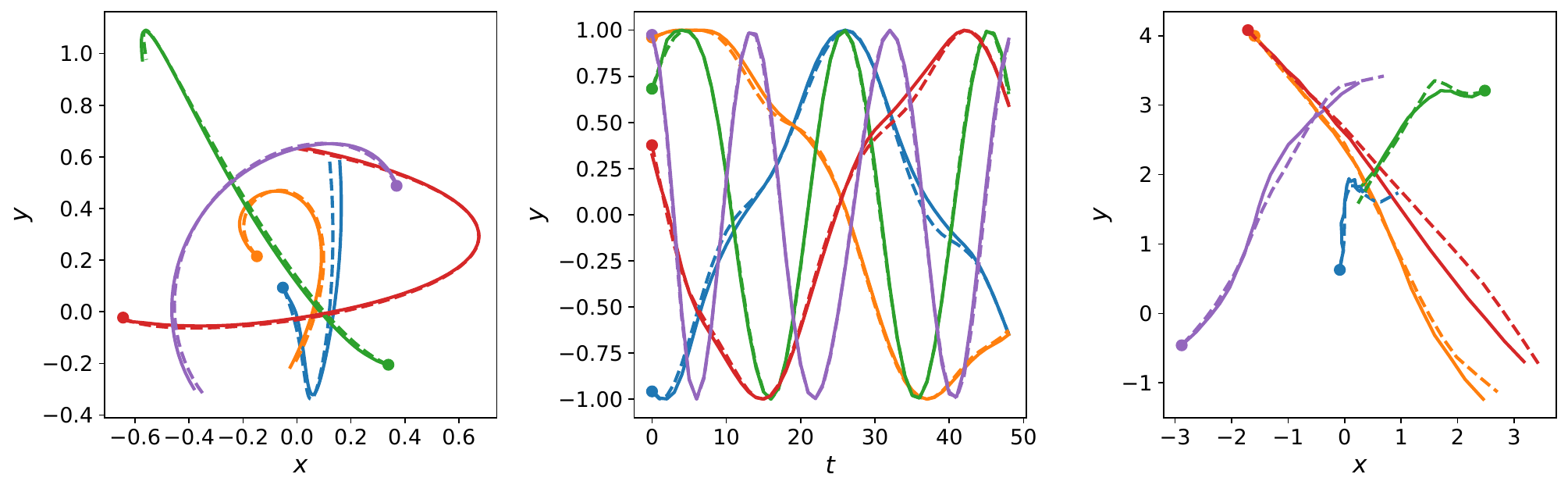}
    \vskip -0.1in
    \caption{\textbf{Trajectory prediction.} We show trajectories predicted by BINN (solid) and the corresponding ground truths (dashed). \textbf{(Left-right)} Examples from mass-spring, Kuramoto oscillator, and TrajNet++ datasets.}
    \label{fig:experiments}
\end{figure*}
In contrast to pendulum preferences, double pendulum preferences are non-mutually exclusive. This observation is supported by the belief matrix, which does not satisfy the mutually-exclusive condition described in Section \ref{subsec:nonlinear_opinion_dynamics},
\begin{equation}
    \mathbf{A}^{\mathrm{o}} = \begin{bmatrix}
        0 & -2.8137\\
        1.1597 & 0
    \end{bmatrix}.
\end{equation}
%
%
\subsection{Trajectory prediction}
We demonstrate the efficacy of our model for trajectory prediction on several datasets. 
For each system, we construct a fully-connected graph with $\mathcal{N}_a$ nodes, and initialize node features to the concatenation of the position and velocity of the corresponding agent. 
We report the quantitative performance of our model
against competitive baselines in Table \ref{tab:experiments}, and show the qualitative performance of our model in Figure~\ref{fig:experiments}. 
Dataset details are provided in Appendix \ref{apx:datasets}.
%
%
\subsubsection{Mechanical systems}
%

\textbf{Mass-spring.} 
We demonstrate the utility of our model for trajectory prediction on the mass-spring dataset. The system consists of five masses and the dynamics are governed by a second-order linear equation.

We compute the communication matrices for our BINN and BINN\raisebox{+0.2ex}{+} models using the inter-agent distance defined in Equation \eqref{eqn:communication_matrix_baseline}.
The predictive performance of our BINN models exceeds the performance of the NRI~\citep{kipf2018neural} and dNRI~\citep{graber2020dynamic} models by an order of magnitude.
%

\textbf{Kuramoto oscillator.} 
The Kuramoto oscillator~\citep{kuramoto1984chemical} dataset describes a 5-agent system with dynamics governed by a first-order nonlinear equation. 

We compute the communication matrices for our BINN and BINN\raisebox{+0.2ex}{+} models using the inter-agent distance defined in Equation~\eqref{eqn:communication_matrix_baseline}.
On this dataset, our BINN models perform comparably with baseline models.
%
%
\subsubsection{Human behavior} \label{subsubsec:experiment_pedestrian_behavior}
%

\textbf{TrajNet++ dataset.} 
The TrajNet++ trajectory forecasting dataset~\citep{kothari2021human} is a collection of interaction-centric multi-agent datasets. We evaluate our model on a synthetic subset of the TrajNet++ dataset. This subset is comprised of 43,697 agent-agent centric trajectories. Each trajectory is $19$-timesteps long and consists of five agents. Since $\Delta t$ is not provided for the synthetic dataset, but is required by our model, we arbitrarily set $\Delta t=1$ for all experiments.
%

We compute the communication matrix for our BINN model using the inter-agent distance defined in Equation~\eqref{eqn:communication_matrix_baseline}. 
For our BINN\raisebox{+0.2ex}{+} model, we assume pedestrian sensing scales inversely with inter-agent distance and introduce a prior $\tilde{\mathbf{A}}^{a}_{(i,j),t}$ on the communication matrix, 
\begin{equation}
    \tilde{\mathbf{A}}^{a}_{(i,j),t} = \frac{1}{\lVert\mathbf{p}_{i,t} - \mathbf{p}_{j,t}\rVert^{2} + \epsilon}.\label{eqn:trajnet_commatrix}
\end{equation}
We augment $\tilde{\mathbf{A}}^{a}_{(i,j),t}$ with a learned multiplier $\mathbf{A}_{(i,j)}^{\mathrm{pre}}$, and form the communication matrix by way of their product, 
\begin{equation}
    \mathbf{A}^{a}_{(i,j),t} = \mathbf{A}_{(i,j)}^{\mathrm{pre}}\odot\tilde{\mathbf{A}}^{a}_{(i,j),t}.\label{eqn:trajnet_augcommatrix}
\end{equation}

\begin{figure}[t]
    \begin{subfigure}{\columnwidth}
        \centering
        \includegraphics[width=\columnwidth]{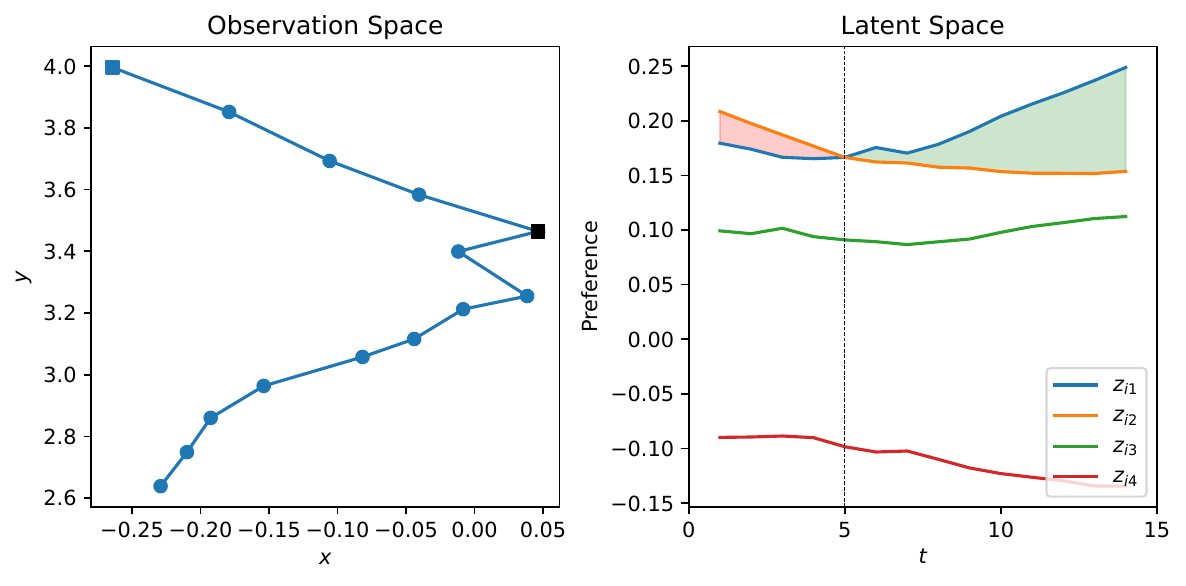}
    \end{subfigure}
    
    \begin{subfigure}{\columnwidth}
        \centering
        \includegraphics[width=\columnwidth]{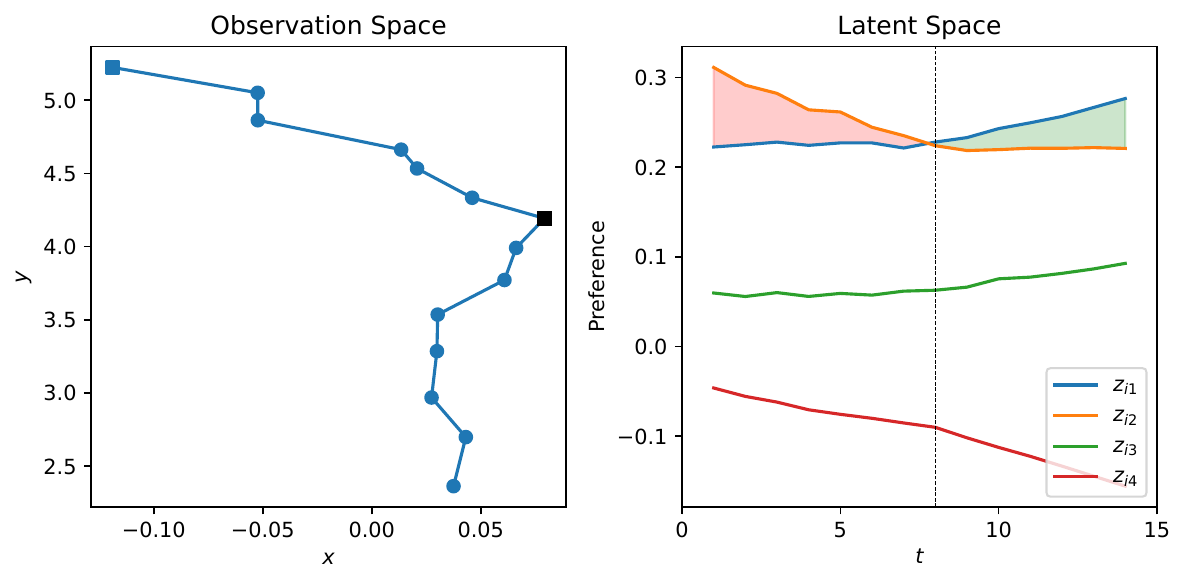}
    \end{subfigure}

    \begin{subfigure}{\columnwidth}
        \centering
        \includegraphics[width=\columnwidth]{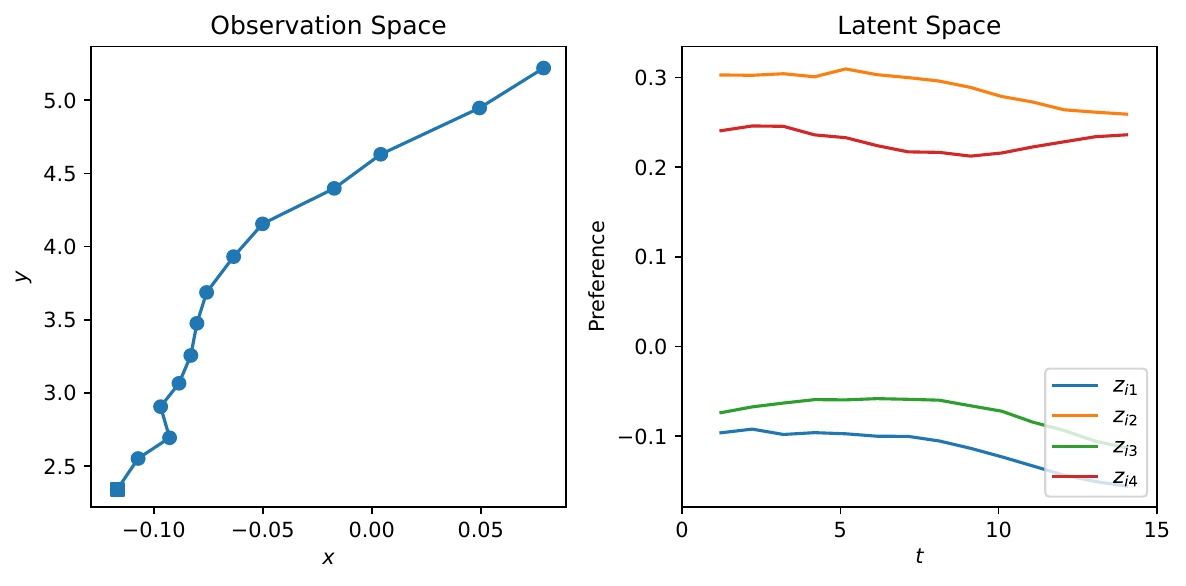}
    \end{subfigure}
    \caption{\textbf{TrajNet++ preferences.} \textbf{(Left)} We visualize example trajectories from the dataset for a single agent. The initial position of the trajectory is marked by a blue square and the point of direction change is marked by a black square. \textbf{(Right)} We plot the agent preferences for each latent categories. Dominant $z_{i1}$ corresponds to leftwards motion and dominant $z_{i2}$ corresponds to rightward motion. Dominant $z_{i3}$ corresponds to downwards motion and dominant $z_{i4}$ corresponds to upwards motion.}
    \label{fig:trajnet_4d}
    \vskip -0.2in
\end{figure}

On this dataset, our BINN and BINN\raisebox{+0.2ex}{+} models outperform all baseline models.
The relationship between the observed trajectories and learned preferences is shown in Figure \ref{fig:trajnet_4d}. 
Three consistent trends emerge across examples. First, when the relative magnitudes of $z_{i1}$ and $z_{i2}$ change, the direction of travel changes from right to left.
Second, when the relative magnitude of $z_{i3}$ to $z_{i4}$ increases, agents increase their y-velocity.
Finally, the dominant $z_{i3}$ in row 1 and 2 results in downwards motion, while the dominant $z_{i4}$ in row 3 results in upwards motion.
%
%

\textbf{NBA player dataset.}
The SportUV dataset~\citep{Linou2016NBA} is comprised of 358,217 real-world NBA player trajectories. Each trajectory is 15 timestep long and consists of 10 agents.

We compute the communication matrix for our BINN model using the inter-agent distance defined in Equations~\eqref{eqn:communication_matrix_baseline}, and the communication matrix for our BINN\raisebox{+0.2ex}{+} model using Equation~\eqref{eqn:trajnet_commatrix} and \eqref{eqn:trajnet_augcommatrix}.
Our BINN$+$ model performs comparably to the dNRI model. 
The relationship between the observed trajectories and learned preferences is shown in Appendix~\ref{apx:experiments_additional_nba}, Figure \ref{fig:nba_4d}.
%
\subsection{Robustness to choice of latent dimension}
We investigate the robustness of our BINN model to the choice of latent dimension. We report the performance of our BINN model for large and small latent dimension in Table~\ref{tab:overparameterization} . Across experiments, the latent dimension of BINN-large is twice that of BINN-small. The performance variation across models is minimal for all experiments, demonstrating the robustness of our model to the choice of latent dimension. We attribute this robustness to the nonlinear opinion dynamics inductive bias which imparts a mechanism for identifying mutually exclusive categories. We report the latent dimension used in each experiment in Appendix~\ref{apx:training_details}, Table~\ref{tab:hyperparameter}. 
%
%
%
\begin{table}[t]
    \renewcommand{\arraystretch}{1.3}
    \centering
    \caption{\textbf{Robustness to choice of latent dimension.} We compare the trajectory prediction performance of our BINN model with different latent dimensions. We use a lower latent dimension in BINN-small and a higher latent dimension in BINN-large. The predictive performance across models is comparable.}
    \setlength\tabcolsep{3pt}
    \begin{tabular}{lccc}
        \hline
        Dimension & Mass-Spring & Kuramoto & TrajNet++\\
        \hline
        BINN-small & $\boldsymbol{2.88\times10^{-4}}$ & $4.98\times10^{-3}$ & $\boldsymbol{1.20\times10^{-2}}$\\
        BINN-large & $3.33\times10^{-4}$ & $\boldsymbol{4.52\times10^{-3}}$ & $2.21\times10^{-2}$\\
        \hline
    \end{tabular}
    \label{tab:overparameterization}
    \vskip -0.1in
\end{table}
%
%
\section{Conclusion}
In this paper, we address the challenge of designing a relational inference model with inter-agent relationships that are informed by preferences for multiple categories. To do this, we propose the use of nonlinear opinion dynamics as an inductive bias, and introduce an interpretable and flexible relational inference model. In addition to outperforming competitive existing relational inference models on multiple trajectory prediction datasets, our model provides a mechanisms for identification of mutually exclusive categories, and controlling preference formation.
%

\textbf{Limitations and future work.} 
Similarly to prior work, our BINN models assume the same agents are present throughout a trajectory.
Efforts to address this limitation would broaden the applicability of this work.
%
%
\subsubsection*{Acknowledgements}
This work was partially supported by NSF grant MRSEC DMR-2011750 and the Princeton Catalysis Initiative.
We sincerely thank the AISTATS 2025 reviewers and area chair for their thoughtful feedback.
%
\clearpage
\bibliographystyle{iclr2025_conference}
\bibliography{ref}
\section*{Checklist}
 \begin{enumerate}
 \item For all models and algorithms presented, check if you include:
 \begin{enumerate}
   \item A clear description of the mathematical setting, assumptions, algorithm, and/or model: \textbf{Yes. Theoretical background for opinion dynamics is described in Section~\ref{sec:background} and Appendix~\ref{apx:nonlinear_opinion_dynamics} and description of the algorithm is given in Section \ref{sec:method} and Figure \ref{fig:BINN_architecture}.}
   \item An analysis of the properties and complexity (time, space, sample size) of any algorithm: \textbf{Yes. Memory requirement for the proposed method and comparison with baselines are given in Table~\ref{tab:computational_expense}.}
   \item (Optional) Anonymized source code, with specification of all dependencies, including external libraries. \textbf{Code is open sourced at \url{https://github.com/CAB-Lab-Princeton/Behavior-Inspired-Neural-Network}.}
 \end{enumerate}

 \item For any theoretical claim, check if you include:
 \begin{enumerate}
   \item Statements of the full set of assumptions of all theoretical results: \textbf{Yes. Assumptions we make about the nonlinear opinion dynamics model are given in Section \ref{sec:background} and Appendix \ref{apx:experiments_additional_mutually_exclusive}.}
   \item Complete proofs of all theoretical results. \textbf{Not applicable. Theoretical results are not part of the contribution to this paper.}
   \item Clear explanations of any assumptions: \textbf{N/A.}
 \end{enumerate}

 \item For all figures and tables that present empirical results, check if you include:
 \begin{enumerate}
   \item The code, data, and instructions needed to reproduce the main experimental results (either in the supplemental material or as a URL): \textbf{Yes. All data generation and training hyper-parameters are given in Table \ref{tab:data_simulation} and \ref{tab:hyperparameter}. Code is open sourced.}
   \item All the training details (e.g., data splits, hyperparameters, how they were chosen): \textbf{Yes. Training details are given in Table \ref{tab:hyperparameter}.}
   \item A clear definition of the specific measure or statistics and error bars (e.g., with respect to the random seed after running experiments multiple times): \textbf{N/A.}
   \item A description of the computing infrastructure used. (e.g., type of GPUs, internal cluster, or cloud provider): \textbf{Yes. Information is provided in Appendix \ref{apx:training_details}. Specific information on the Neuronic cluster is availible at \url{https://clusters.cs.princeton.edu}.}
 \end{enumerate}

 \item If you are using existing assets (e.g., code, data, models) or curating/releasing new assets, check if you include:
 \begin{enumerate}
   \item Citations of the creator If your work uses existing assets: \textbf{Yes. TrajNet++~\citep{kothari2021human} and SportsVU~\citep{Linou2016NBA} are cited.}
   \item The license information of the assets, if applicable: \textbf{N/A.}
   \item New assets either in the supplemental material or as a URL, if applicable: \textbf{N/A.}
   \item Information about consent from data providers/curators: \textbf{N/A.}
   \item Discussion of sensible content if applicable, e.g., personally identifiable information or offensive content: \textbf{N/A.}
 \end{enumerate}

 \item If you used crowdsourcing or conducted research with human subjects, check if you include:
 \begin{enumerate}
   \item The full text of instructions given to participants and screenshots: \textbf{N/A.}
   \item Descriptions of potential participant risks, with links to Institutional Review Board (IRB) approvals if applicable: \textbf{N/A.}
   \item The estimated hourly wage paid to participants and the total amount spent on participant compensation: \textbf{N/A.}
 \end{enumerate}

 \end{enumerate}

%
\onecolumn
\section*{Appendix}
\renewcommand{\thesubsection}{\Alph{subsection}}
\setcounter{subsection}{0}
%
%
\subsection{Datasets}\label{apx:datasets}
%
%
\subsubsection{Mechanical systems}
All simulated mechanical system datasets were generated similarly. 
For each system, we generate 50,000 training trajectories, and 12,500 validation and testing trajectories. 
We note the numerical integrator, timestep, trajectory length and system parameters used to simulate each system in Table \ref{tab:data_simulation}. 
We coarsen each simulated trajectory to reduce dataset size, and training time; the coarsening frequency for each system is provided in Table \ref{tab:data_simulation}.
\begin{table}[b]
    \centering
    \caption{\textbf{Data generation and post processing parameters.}}
    \begin{tabular}{lccccc}
        \hline
        & Pendulum & Double Pendulum & Mass-Spring & Kuramoto & TrajNet++\\
        \hline
        Integrator & RK4 & RK4 & RK4 & RK4 & n/a\\
        $\Delta t$ & $1\text{e-}3$ & $5\text{e-}4$ & $5\text{e-}4$ & $5\text{e-}4$ & $1.0$ \\
        Steps & $5000$ & $5000$ & $5000$ & $500$ & n/a \\
        Coarsening & $100$ & $100$ & $100$ & $10$ & n/a \\
        \hline
    \end{tabular}
    \label{tab:data_simulation}
    \vskip -0.in
\end{table}

\textbf{Pendulum.}
We generate a synthetic dataset of pendulum motion, using the second-order dynamical equation,
\begin{equation}
    \ddot{\theta} = -\frac{g}{\ell}\sin\theta,
\end{equation}
with $\ell=1.0$ and gravity $g=9.81$. 
For each trajectory, we sample initial conditions from the uniform distribution on the interval $\left[-0.5\pi, 0.5\pi\right]$. 
%

\textbf{Double pendulum.}
We generate a synthetic dataset of double pendulum motion, using the second-order dynamical equation,
\begin{align}
    \left(m_1+m_2\right) l_1 \ddot{\theta}_1+m_2 l_2 \ddot{\theta}_2 \cos \left(\theta_2-\theta_1\right) & =m_2 l_2 \dot{\theta}_2^2 \sin \left(\theta_2-\theta_1\right)-\left(m_1+m_2\right) g \sin \theta_1, \\
    l_2 \ddot{\theta}_2+l_1 \ddot{\theta}_1 \cos \left(\theta_2-\theta_1\right) & =-l_1 \dot{\theta}_1^2 \sin \left(\theta_2-\theta_1\right)-g \sin \theta_2,
\end{align}
with $\ell_{1}=\ell_{2}=1.0$, $m_{1}=m_{2}=1.0$, and gravity $g=9.81$.
For each trajectory, we sample initial conditions from the uniform distribution on the interval $\left[-0.5\pi, 0.5\pi\right]$. 
%

\textbf{Mass-spring.}
We generate a synthetic dataset of mass-spring motion, using the second-order linear dynamical equation,
\begin{equation}
    \ddot{\mathbf{r}}_{i} = \sum_{i\neq j}^{\mathcal{N}_{a}} -k_{ij}\left(\mathbf{r}_{i}-\mathbf{r}_{j}\right),
\end{equation}
where the spring constant $k_{ij}$ is either $2.5$ or $0$, and $k_{ij}=k_{ji}$. 
The initial conditions are sampled from a normal distribution with $\mu=0$ and $\sigma=0.3$.
%

\textbf{Kuramoto oscillator.}
We generate a synthetic dataset of Kuramoto oscillator~\citep{kuramoto1984chemical} motion where the oscillator dynamics are defined by the first-order nonlinear equation,
\begin{equation}
        \dot{\phi}_{i} = \omega_{i} + \sum_{i\neq j} k_{ij}\sin\left(\phi_{i}-\phi_{j}\right),
\end{equation}
where the coupling constant $k_{ij}$ is either $2.5$ or $0$ and $k_{ij}=k_{ji}$. 
The initial conditions are sampled from a normal distribution with $\mu=0$ and $\sigma=2\pi$.
%
%
\subsubsection{Human behavior}
%

\textbf{TrajNet++ dataset.} 
The TrajNet++ dataset~\cite{kothari2021human} is divided into real and synthetic pedestrian trajectory datasets; we use the synthetic dataset in our experiments. The synthetic dataset is comprised of $43,697$ trajectories, and each trajectory has $19$ timesteps and five agents. 
%

\textbf{NBA player movement dataset.}
The SportsVU dataset~\cite{Linou2016NBA} consists of NBA player trajectories captured over 631 games in the 2015/16 season. 
The dataset consists of $358,217$ trajectories, and each trajectory has $15$ timesteps and $11$ agents ($10$ players and the ball). 
We train our model on player trajectories and omit the ball.
%
%
\subsection{Training Details}\label{apx:training_details}
In this section, we provide training details for each experiment.
Our source code is publicly available at \url{https://github.com/CAB-Lab-Princeton/Behavior-Inspired-Neural-Network}.

All models were trained on Dell Precision 7920 work stations. Each work station is equipped with an Intel Xeon Gold 5220R 24 core CPU, two Nvidia A6000 GPUs, and 256GB of RAM.

We trained each model on a single GPU with CPU and RAM shared between two workloads.
We report memory usage in Table \ref{tab:computational_expense} and training hyperparameters in Table \ref{tab:hyperparameter}. Memory usage is measured at the end of the first training epoch using NVIDIA-smi. 
\begin{table}[t]
    \renewcommand{\arraystretch}{1.3}
    \centering
    \caption{\textbf{Computational cost.} Memory usage for BINN (ours), NRI, and dNRI is reported. Our model (BINN) has the lowest memory usage for all datasets (lower is better).}
    \setlength\tabcolsep{5pt}
    \begin{tabular}{lccc}
        \hline
        Network & Mass-Spring & Kuramoto & TrajNet++\\
        \hline
        BINN* & $\mathbf{496}$ MiB & $\mathbf{484}$ MiB & $\mathbf{406}$ MiB\\
        NRI & $1670$ MiB & $1658$ MiB & $916$ MiB\\
        dNRI & $5008$ MiB & $4996$ MiB & $2180$ MiB\\
        \hline
    \end{tabular}
    \label{tab:computational_expense}
    \vskip -0.1in
\end{table}
\begin{table}[b]
    \centering
    \caption{\textbf{Experiment hyperparameters.}}
    \begin{tabular}{lccccc}
        \hline
        Hyper-parameter & Pendulum & Double Pendulum & Mass-Spring & Kuramoto & TrajNet++\\
        \hline
        $\mathcal{N}_{\mathrm{a}}$ & $1$ & $2$ & $5$ & $5$ & $5$\\
        $\mathcal{N}_{\mathrm{o}}$ & $2$ & $2$ & $4$ & $2$ & $4$\\
        Epoch & $500$ & $500$ & $1000$ & $1000$ & $1000$ \\
        Batch Size & $256$ & $256$ & $256$ & $256$ & $256$\\
        Activation & ReLU & ELU & tanh & tanh & tanh\\
        Optimizer & Adam & Adam & Adam & Adam & Adam\\
        Initial Learning Rate & $1\text{e-}3$ & $1\text{e-}3$ & $1\text{e-}3$ & $1\text{e-}3$ & $5\text{e-}4$\\
        Hidden Dimension & $64$ & $64$ & $128$ & $128$ & $128$\\
        Scheduler & n/a & n/a & StepLR & StepLR & StepLR\\
        Scheduler Step & n/a & n/a & $200$ & $200$ & $200$\\
        Scheduler Gamma & n/a & n/a & $0.25$ & $0.25$ & $0.25$\\
        $\gamma_{1}$ & 1.0 & 1.0 & 1.0 & 1.0 & 1.0\\
        $\gamma_{2}$ & 1.0 & 1.0 & 1.0 & 1.0 & 1.0\\
        \hline
    \end{tabular}
    \label{tab:hyperparameter}
    \vskip -0.1in
\end{table}
%
%
\subsection{Additional Experimental Results}\label{apx:experiments_additional}
%
%
\subsubsection{Consistency across random seeds}\label{apx:experiments_additional_double_pendulum}

\begin{figure}[t]
    \begin{subfigure}{0.5\columnwidth}
        \centering
        \includegraphics[width=0.96\textwidth]{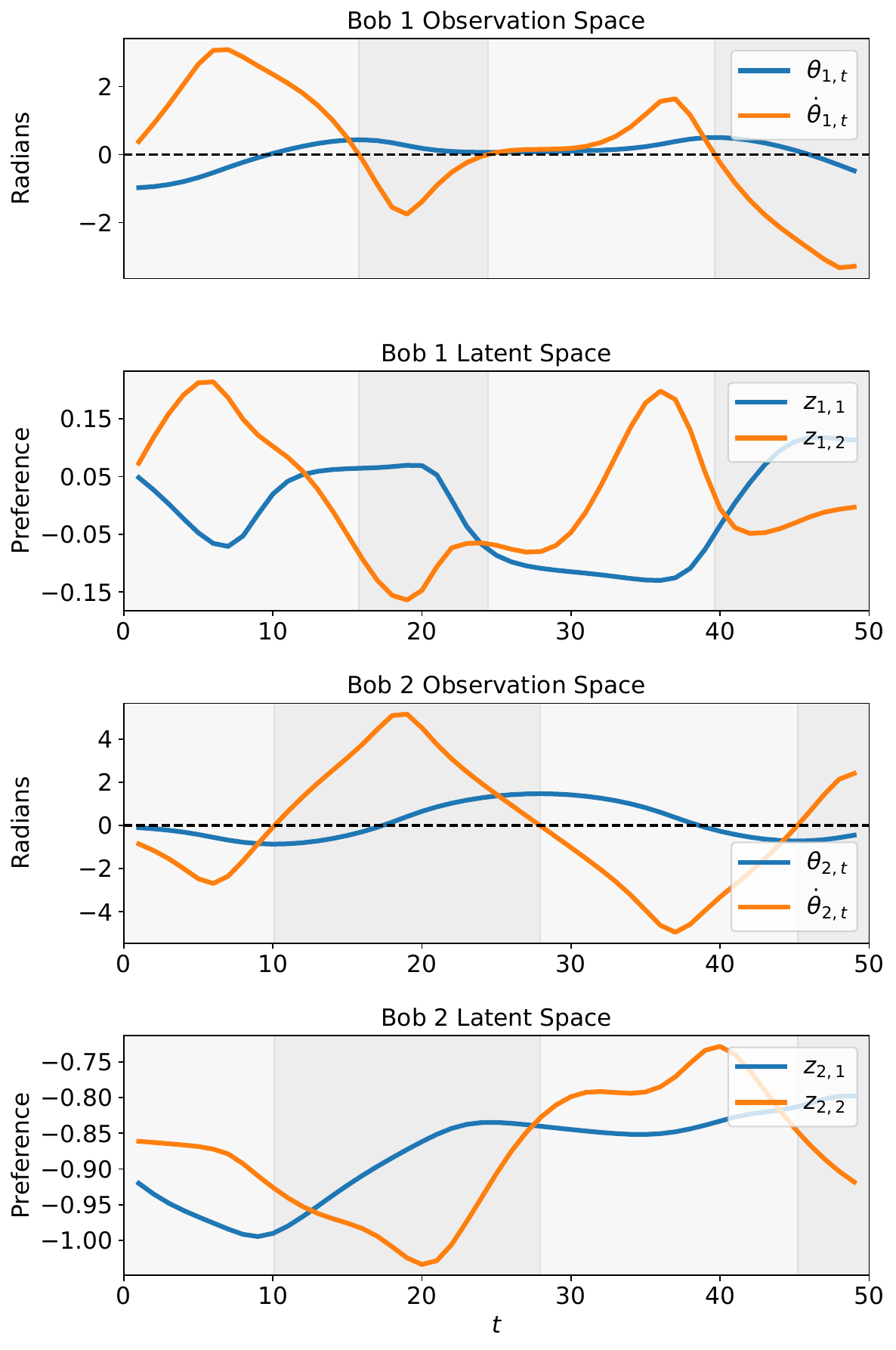}
    \end{subfigure}
    \begin{subfigure}{0.5\columnwidth}
        \centering
        \includegraphics[width=0.96\textwidth]{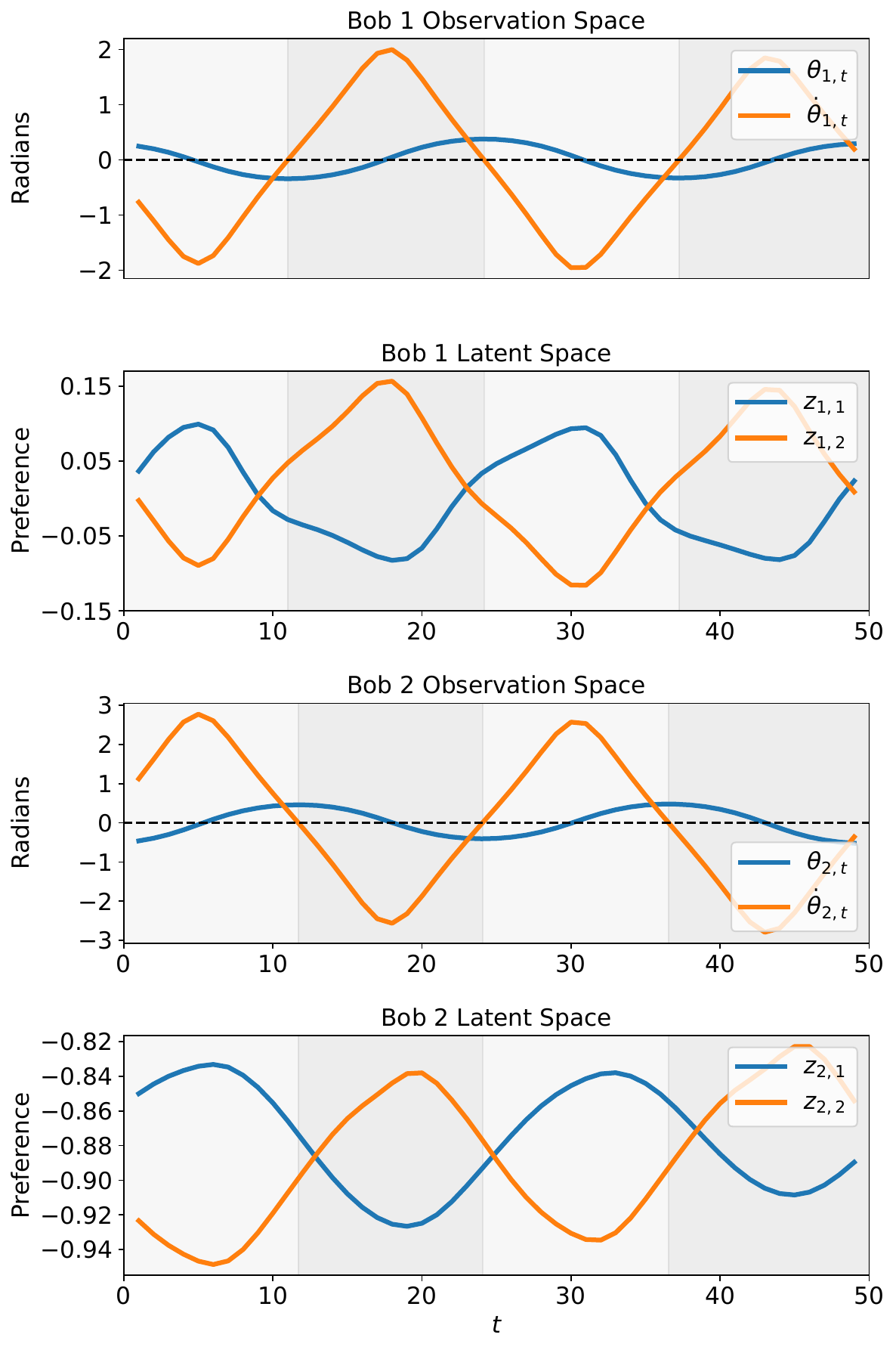}
    \end{subfigure}
    \caption{\textbf{Consistency across random seeds.} We show the observed, and latent representations of each double pendulum bob. \textbf{(Left-right)} Encodings learned with a random seed of 172; encodings learned with a random seed of 272. \textbf{(Top-bottom)} The observed position and velocity of the first bob; the learned preferences of the first bob; the observed position and velocity of the second bob; the learned preferences of the second bob. When preference switch, physical bob's motion switches direction.}
    \label{fig:double_pendulum_2d_1}
    \vskip -0.1in
\end{figure}

The character of the preferences learned in our BINN model is consistent across random seeds. In Figure~\ref{fig:double_pendulum_2d_1}, we provide supplemental observation space to preference encodings for the double pendulum system presented in Section~\ref{sec:experiment}. The encodings on the left were learned with a random seed of 172, and the encodings on the right were learned with a random seed of 272. The double pendulum encodings in Figure~\ref{fig:double_pendulum_2d} were learned with a random seed of 72.
%
%
\subsubsection{Capturing real-world human behavior}\label{apx:experiments_additional_nba}
\begin{figure}[t]
    \begin{subfigure}{\columnwidth}
        \centering
        \includegraphics[width=0.5\textwidth]{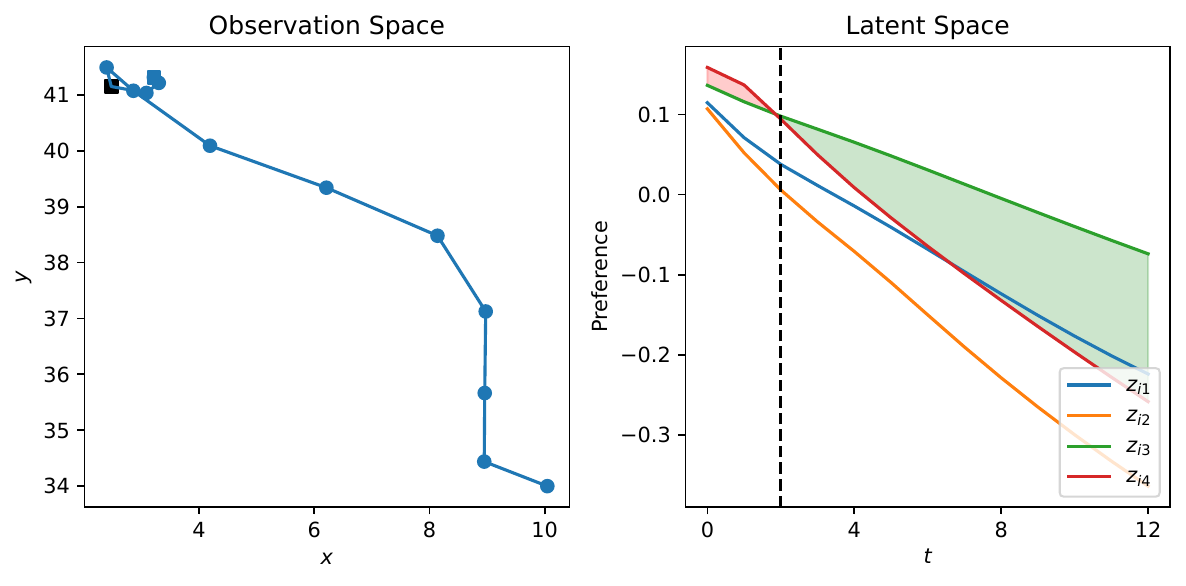}
    \end{subfigure}
    \begin{subfigure}{\columnwidth}
        \centering
        \includegraphics[width=0.5\textwidth]{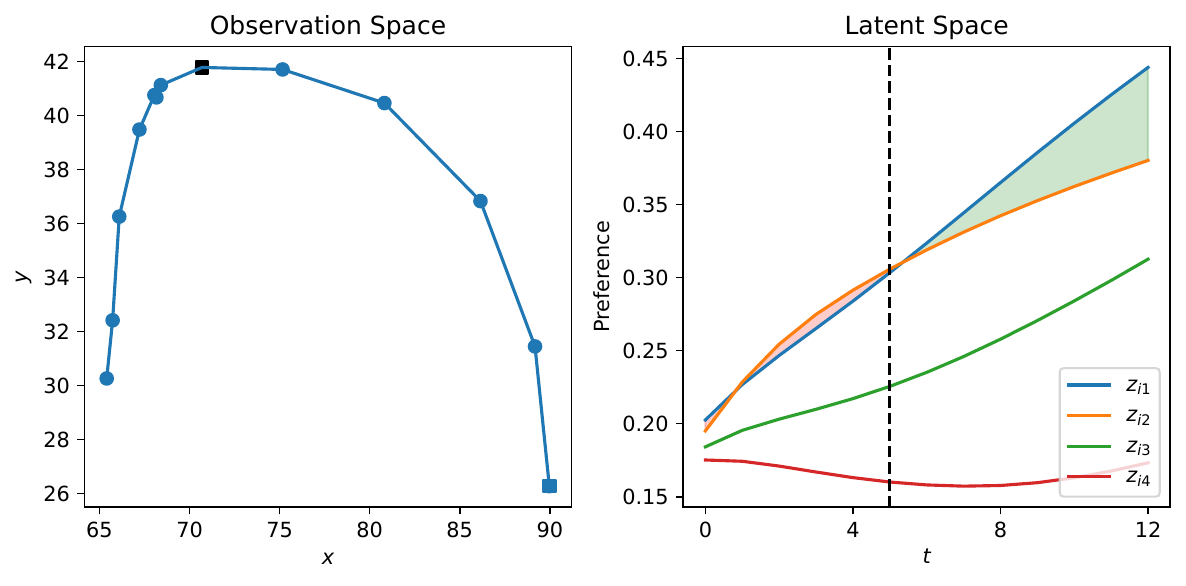}
    \end{subfigure}
    \caption{\textbf{NBA player preferences.} \textbf{(Left)} We visualize example trajectories from the dataset. The initial position of the trajectory is marked by a blue square and the point of direction change is marked by a black square. \textbf{(Right)} We plot agent preferences for each latent categories. Dominant $z_{i1}$ corresponds to upward motion, and dominant $z_{i2}$ corresponds to downward motion. Dominant $z_{i3}$ corresponds to rightward motion, and dominant $z_{i4}$ corresponds to leftward motion.}
    \label{fig:nba_4d}
    \vskip -0.1in
\end{figure}

Human behavior can be less predictable than simulated systems. To assess how well our model can capture the behavior of human agents, we performed additional experiments on the SportsVU dataset~\cite{Linou2016NBA}. We provide observation space to preference encodings for the SportsVU dataset~\cite{Linou2016NBA} in Figure~\ref{fig:nba_4d}. The learned relationship between preferences is similar to the learned relationship for synthetic pedestrian data (see Section~\ref{subsubsec:experiment_pedestrian_behavior}).
%
%
When the relative magnitudes of $z_{i1}$ and $z_{i2}$ change, the direction of travel changes from upwards to downwards.
When the relative magnitudes of $z_{i3}$ and $z_{i4}$ change, the direction of travel changes from leftwards to rightwards.
%
%
\subsubsection{Additional baseline comparisons}
We report an additional comparison against Factorised Neural Relational Inference (fNRI)~\citep{webb2019factorised}. Similarly to our approach, fNRI models multiple types of inter-agent relationships; however, their approach uses a factorized graph representation, while ours uses the nonlinear opinion dynamics inductive bias.
We report the 10-step trajectory prediction performance of our BINN\raisebox{+0.24ex}{+} 
and fNRI on the TrajNet++ dataset in Table~\ref{tab:experiment_fnri}, where our BINN\raisebox{+0.24ex}{+} model outperforms fNRI.
%
%
\begin{table}[b]
    \renewcommand{\arraystretch}{1.3}
    \setlength{\tabcolsep}{10pt}
    \centering
    \caption{\textbf{Trajectory prediction.} Trajectory prediction error on the TrajNet++ datasets is reported. Our BINN\raisebox{+0.2ex}{+} model outperforms fNRI.}
    \begin{tabular}{lccc}
        \hline
        Network & TrajNet++\\
        \hline
        BINN\raisebox{+0.24ex}{+}* & $\boldsymbol{3.51\times10^{-3}}$\\
        fNRI & $6.85\times10^{-3}$\\
        \hline
    \end{tabular}
    \label{tab:experiment_fnri}
\end{table}
%
%
\subsection{Aspects of Bifurcation Theory and Nonlinear Opinion Dynamics }\label{apx:nonlinear_opinion_dynamics}

In this section, we aim provide intuition building examples for the pitchfork bifurcation introduced in Figure~\ref{fig:nonlinear_opinion_dynamics_bifurcation}, and the notion of mutually exclusive categories introduced in Section~\ref{subsec:nonlinear_opinion_dynamics}.

\textbf{Pitchfork bifurcation.}
Bifurcation theory can be described as the study of equations with multiple solutions. Consider, for example, the dynamical system defined by
\begin{equation}
    \dot{z} = z^{3} - uz.\label{eqn:pitchfork_example}
\end{equation}
The equilibrium solutions of this dynamical equation (i.e., the values of $z$ for which $\dot{z}=0$) are shown in the bifurcation diagram of Figure~\ref{fig:pitchfork_example}. 
When the bifurcation parameter $u$ is less than or equal to the critical value $u^* = 0$, there is only one solution for Equation~\ref{eqn:pitchfork_example} (i.e., $z=0$). 
When $u$ is greater than $u^*$, the number of solutions jumps to three (i.e., $z\in\{\sqrt{u}, -\sqrt{u}, 0\}$). 
The jump in the number of solutions from one to three at the critical value $u^*$, is the basic property that characterizes a pitchfork bifurcation. As it turns out, Equation~\ref{eqn:pitchfork_example} is the simplest equation with this behavior~\citep{golubitsky2012singularities}.

The stability of solutions to the dynamical system defined in Equation~\eqref{eqn:pitchfork_example} can be determined by evaluating the derivative
\begin{equation}
    \ddot{z} = 3z^{2} - u.
\end{equation}
When the derivative is greater than zero, the solution is stable. 
When it is less than zero, the solution is unstable. For the solution, $z=0$, the derivative evaluates to
\begin{equation}
    \ddot{z} = -u,
\end{equation}
which is stable for $u < 0$, and unstable for $u > 0$. The change in the stability of the solution $z=0$ is illustrated in the bifurcation diagram of Figure~\ref{fig:pitchfork_example}. When $u<0$, the solution $z=0$ is illustrated with a solid line and when $u>0$, the solution is illustrated with a dotted line. The stable solutions $z\in\{\sqrt{u}, -\sqrt{u}\}$ are illustrated with solid lines.
\begin{figure}[t]
    \centering
    \resizebox{0.4\textwidth}{!}{
        \begin{tikzpicture}[>={stealth'}, line width = 0.25mm]
        \draw[->] (-13,-1)--(-9,-1) node [right] {\tiny $u_{i}$};
        \draw (-11,-1)--(-11,-0.9);
        \node [below] at (-11,-1) {\tiny $u^{*}$};
        \draw[->] (-13,-1)--(-13,1) node [left] {\tiny $z_{ij}$};
        \draw (-13,0)--(-12.9,0);
        \node [left] at (-13,0) {\tiny $0$};
        \draw (-13,0)--(-11,0);
        \draw[dotted] (-11,0)--(-9,0);
        \draw[color=black, domain=-11:-9, samples=150] plot (\x, {0.07*sqrt(50*(\x+11))});
        \draw[color=black, domain=-11:-9, samples=150] plot (\x, {-0.07*sqrt(50*(\x+11))});
        \end{tikzpicture}}
    \caption{\textbf{Pitchfork bifurcation.} States are represented by $z$ and $u$ is the bifurcation parameter. When $u$ is less than the critical value $u^{*}$, there is only one solution, $z=0$. When $u$ is larger than the critical value $u^{*}$, the solutions changes to three, $z\in\{\sqrt{u},-\sqrt{u},0\}$. The stable (unstable) equilibrium solutions are illustrated with solid (dotted) lines. }
    \label{fig:pitchfork_example}
    \vskip -0.1in
\end{figure}

In applications, the bifurcation parameter often corresponds to an external force. An example of a physical system whose solutions form a pitchfork bifurcation is a trailer being towed by a truck.
In this case, the location of the center of gravity of the trailer is the bifurcation parameter and determines the stability of the truck-trailer system. 
A video showing the stable and unstable solutions of this physical system can be found at \url{https://www.youtube.com/watch?v=w9Dgxe584Ss}. 
A longer discussion of this system can be found in~\cite{horvath2022stability}.
%

\textbf{Separability of mutually exclusive categories.} \label{apx:experiments_additional_mutually_exclusive}
In the nonlinear opinion dynamics framework, the presence of mutually exclusive preferences simplifies the dynamics. 
Consider the case of two categories (i.e., $\mathcal{N}_{\mathrm{o}} = 2$). 
Preferences that are opposite in sign (i.e., $z_{i1}=-cz_{i2}$) and negatively correlated (i.e., $a^{\mathrm{o}}_{12}, a^{\mathrm{o}}_{21}<0$) are mutually exclusive. 
Given these conditions, the belief term can be simplified to
\begin{equation}
    \sum_{\substack{l=1\\ l\neq j}}^{\mathcal{N}_{\mathrm{o}}}a^{\mathrm{o}}_{jl}z_{il} = a_{21}z_{i2} = -ca_{12}z_{i1},
\end{equation}
and the communication-belief term can be simplified to
\begin{equation}
    \sum_{\substack{k=1\\ k\neq i}}^{\mathcal{N}_{\mathrm{a}}}\sum_{\substack{l=1\\ l\neq j}}^{\mathcal{N}_{\mathrm{o}}} a^{\mathrm{a}}_{ik}a^{\mathrm{o}}_{jl}z_{kl} = \sum_{\substack{k=1\\ k\neq i}}^{\mathcal{N}_{\mathrm{a}}} a^{\mathrm{a}}_{ik}a^{\mathrm{o}}_{21}z_{k2} = \sum_{\substack{k=1\\ k\neq i}}^{\mathcal{N}_{\mathrm{a}}} -a^{\mathrm{a}}_{ik}ca^{\mathrm{o}}_{12}z_{k1}.
\end{equation}
With these simplification, the dynamics of $z_{i1}$ in  Equation~\eqref{eqn:nonlinear_opinion_dynamics} do not depend on $z_{i2}$.
Defining the decoupled self-reinforcement term
\begin{equation}
    \tilde{\alpha}_{i} = \alpha_{i1} - ca^{o}_{12},
\end{equation}
and communication matrix
\begin{align}
    \tilde{a}_{ik} &= a^{a}_{i1} - ca^{a}_{i1}a^{0}_{12},
\end{align}
Equation \eqref{eqn:nonlinear_opinion_dynamics} can be simplified to Equation~\eqref{eqn:nonlinear_opinion_dynamics_mutually_exclusive_reduced}, that is,
\begin{equation}
    \dot{z}_{i} = -d_{i}z_{i} + \mathcal{S}\left(u_{i}\left(\tilde{\alpha}_{i}z_{i} + \sum_{\substack{k=1\\ k\neq i}}^{\mathcal{N}_{\mathrm{a}}}\tilde{a}_{ik}z_{k}\right)\right) + b_{i}.
\end{equation}

\end{document}